\documentclass[runningheads]{llncs}

\usepackage{latexsym}
\usepackage{graphicx}
\usepackage{varioref}
\usepackage{url}
\usepackage[utf8]{inputenc}
\usepackage{epsfig}
\usepackage{amsmath}
\usepackage{amssymb}
\usepackage{graphicx}
\usepackage{algorithm2e}
\usepackage{algorithmic}
\usepackage{subfigure}
\usepackage{multirow}
\usepackage{float}

\def\urltilda{\kern -.15em\lower .7ex\hbox{\~{}}\kern .04em}
\newtheorem{mydefinition}[definition]{Definition}
\newtheorem{myexample}[definition]{Example}
\newtheorem{myremark}[definition]{Remark}

\begin{document}

\pagestyle{headings}

\mainmatter

\title{Logical Explanations for Deep Relational Machines Using Relevance Information}
\author{Ashwin Srinivasan\inst{1} \and Lovekesh Vig\inst{2} \and Michael Bain\inst{3}}
\institute{Department of Computer Sc. \& Information Systems \\
        BITS Pilani, K.K. Birla Goa Campus, Goa \\ \and
        TCS Research, New Delhi.\and
        School of Computer Science and Engineering \\
        University of New South Wales, Sydney, NSW.}

\maketitle
\begin{abstract}
Our interest in this paper is in the construction of symbolic explanations
for predictions made by a deep neural network. We will focus attention
on deep relational machines (DRMs: \cite{lodhi:drm}).
A DRM is a deep network in which the input layer consists of Boolean-valued functions (features)
that are defined in terms of relations provided as domain, or background, knowledge. Our
DRMs differ from those in \cite{lodhi:drm}, which uses an Inductive Logic
Programming (ILP) engine to first select features (we use random selections from
a space of features that satisfies some approximate
constraints on logical relevance and non-redundancy). But why do the
DRMs predict what they do? One way of answering this was provided in recent work \cite{lime}, by constructing readable proxies for a black-box predictor. The proxies are intended only to model the predictions of the black-box in local regions of the instance-space. But readability alone may not enough: to be understandable, the local models must use relevant concepts in an meaningful manner.
We investigate the use of a Bayes-like approach to identify logical proxies for
local predictions of a DRM.
We show: (a) DRM's with our randomised propositionalization method
achieve state-of-the-art predictive performance; (b) Models in first-order logic can approximate the DRM's prediction closely in a small local region; and (c) Expert-provided relevance information can play the role of a prior to distinguish between logical explanations that perform equivalently
on prediction alone.
\end{abstract}

\section{Introduction}
\label{sec:intro}

In \cite{anty}, a contrast is presented between theories that
predict everything, but explain nothing; and those that explain everything, but
predict nothing. Both are seen as having limited value in the scientific enterprise,
which requires models with both predictive and explanatory power.
Michie \cite{dm:window} adds a further twist to this by suggesting that
the limitations of the human brain may force a ``window'' on the complexity
of explanations that can be feasibly understood, even if they are described in some readable form like a natural or a symbolic formal language.

Complexity limits notwithstanding, it is often assumed that that
predictive and explanatory assessments refer the {\em same\/} model.
But this does not have to be so: early results in the Machine Learning
(ML) literature on behavioural cloning point to areas where people perform
predictive tasks using sub-symbolic models, for which entirely separate
symbolic explanations were constructed by machine \cite{michie:superart}.
It is not therefore inevitable that machine learning models be either
purely sub-symbolic or purely symbolic.
 
There are, of course, problems for which one or the other is much better
suited.
Prominent recent examples of successful sub-symbolic learning abound with
the use of deep neural networks in automatic caption learning (``woman in
a white dress standing with tennis racquet'': \cite{karpathy:caption}),
speech-recognition (Siri, Google Now) and machine translation (Google
Translate).
It is equally evident that for learning recursive theories, program
synthesis, and learning with complex domain-knowledge, symbolic techniques
like Inductive Logic Programming (ILP) have proved to be remarkably effective.
But there are also a number of problems that could benefit from an
approach that requires not one or the other, but both forms of learning
(in his book ``Words and Rules'', Steven Pinker conjectures language
learning as one such task).
 
\begin{figure*}[htb]
\centerline{\includegraphics[width=\textwidth,height=0.36\textheight]{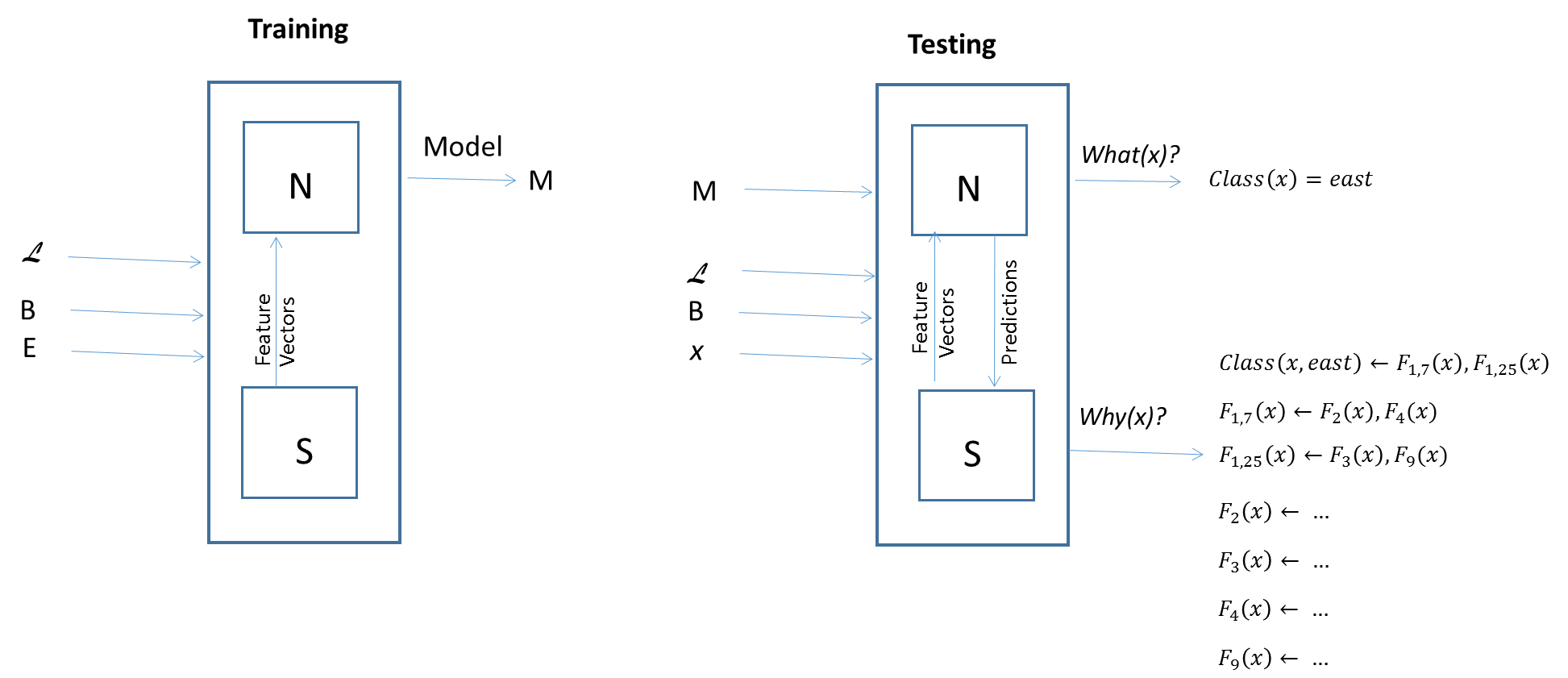}}
\caption{What this paper is about. $N$ is a deep neural network, and $S$ is a symbolic learner.
$B$ is domain-specific background knowledge; ${\cal L}$ language constraints; $E$ a set of
training examples. $N$ is used to construct a predictive model $M$. Explanations
of why $M$ predicts the class of a test instance $x$ are to be obtained as a logical model
that uses features, that are pre-defined in $B$ or ``invented''. 
Correctly, $B$ and ${\cal L}$ on the right are augmentations of those on the left; and the $S$'s on right and left
are different instantiations of a first-order symbolic learner that is capable of producing
definitions of features (left), or explanations (right).}
\label{fig:bigpic}
\end{figure*}

Our interest here therefore is in investigating the use of separate models
for prediction and explanation.
Specifically, we examine the use of a neural model for prediction, and a
logical model for explanation.
But in order to ensure that the models are about the same kinds of things,
the models are not constructed independent of each other.
The neural model is constructed in terms of features in (first-order)
logic identified by a simple form of propositionalization methods
developed in ILP.
In turn, the predictions of the neural model are used to construct
logical models in terms of the features.\footnote{We restrict
ourselves in this paper to queries of the form: ``What is the class of
instance $x$?'', but the ideas extend straightforwardly to tasks other
than classification.}
Figure \ref{fig:bigpic} shows how the various pieces we have just
described fit together.

Having two separate models would appear to be an extravagance that could
be ill-afforded if the neural model wasn't a very good predictor, or if
the logical model wasn't a very good explainer.
In this paper, we present empirical evidence that neither of these appears
to happen here.
For ML practitioners, the principal points of interest in the paper are these:

\begin{enumerate}
\item[(a)] We present a simple randomised method for obtaining first-order features that use domain-knowledge
    for use by a deep neural network. The resulting {\em deep relational machine\/}, or DRM,
    achieves high predictive performance, with small numbers of data instances. This extends
    existing work on DRMs, which have in the past required the use of a full-fledged ILP engine
    for feature-selection \cite{lodhi:drm}; and
\item[(b)] We present a method for extracting local symbolic explanations for predictions made
	by the DRM by employing a Bayes-like trade-off of likelihood and prior preference. 
    The latter is obtained
    using domain-knowledge of relevance of predicates used in the explanations. This extends
    the techniques proposed in \cite{lime} in the direction of making explanations both more
    expressive (by using first-order logic) and meaningful (by using
    a Bayes-like method of incorporating domain-specific preferences into explanations).
\end{enumerate}

The rest of the paper is organised as follows. In Section \ref{sec:drm} we
describe the DRMs we use for prediction.
Section \ref{sec:expl} describes the notion of explanations as used in
this paper. Selection amongst several possible explanations is in Section
\ref{sec:sel}, which introduces the use of a relevance-based prior in
Section \ref{sec:prior}. Section \ref{sec:expt} presents an empirical
evaluation of the predictive and explanatory models, using some benchmark
datasets. Appendix \ref{appa} contains details of the domain-specific
relevance information used in the experiments.

\section{A Deep Relational Machine for Prediction}
\label{sec:drm}

One of the most remarkable recent advances in the area of Machine Learning is a resurgence of
interest in neural networks, resulting from the automated construction of ``deep networks'' for prediction.
Simplistically, Deep Learning is using neural networks with multiple hidden layers.
Mathematically speaking, it is a composition of multiple simple non-linear functions trying to learn a
hierarchy of intermediate features that most effectively aid the global learning task. Learning such
intermediate features with neural networks has been made possible by three separate advances:
(a) mathematical techniques that allow the training of neural networks with very large numbers of hidden layers;
(b) the availability of very large amounts of data that allow the estimation of parameters (weights) in
such complex networks; and (c) the advanced computational capabilities offered by modern day GPUs to train deep
models. Despite successes across a wide set of problems, deep learning is unlikely to be sufficient for
all kinds of data analysis problems.
The principal difficulties appear to lie in the data and computational requirements to train such models.
This is especially the case if many hidden layers are needed to encode complex concepts (features).
For many industrial processes, acquiring data can incur significant costs, and simulators can be computationally very intensive.

Some of this difficulty may be alleviated if knowledge already available in the area of interest
can be taken into account. Consider, for example, a problem in the area of drug-design. Much may be known already about the target of interest: small molecules that have proved to be effective, what can and cannot be synthesized cheaply, and so on.
If these concepts are relevant to constructing a model for predicting good drugs, it seems both unnecessary and inefficient to require a deep network to re-discover them (the problem is actually worse --- it may not even be possible to discover the concepts from first-principles using the data available).
It is therefore of significant practical interest to explore ways in which prior domain-knowledge could be used in deep networks to reduce data and computational requirements.

Deep Relational Machines, or DRMs, proposed in \cite{lodhi:drm}, are
deep neural networks with first-order Boolean functions at the
input layer (``function $F_1$ is true
if the instance $x$ is a molecule containing a 7-membered ring connected to a
lactone ring'' --- definitions of
relations like 7-membered and lactone rings are expected to be present in
the background knowledge). In \cite{lodhi:drm}
the functions are learned by an ILP engine. This follows a long
line of research, sometimes called {\em propositionalization\/},
in which features constructed by ILP have been
used by other learning methods like regression, decision-trees,
SVMs, topic-models, and multiplicative weight-update linear threshold
models. In each of these, the final model is constructed in two
steps: first, an ILP engine constructs a set of ``good'' features, and
then, the final model is constructed using these features, possibly
in conjunction with other features already available. Usually
the models show significant improvements in predictive performance when
an existing feature set is enriched in this manner. In \cite{lodhi:drm},
the deep network with ILP-features is shown to perform well, although
the empirical evidence is limited. For the DRMs used in this paper we dispense
with the requirement for an ILP-based selection of input features.
Instead we use random selections of features generated from a space that has some constraints on logical relevance and redundancy. 
We will continue to refer to this kind of model as a DRM, since
the  features are still first-order Boolean functions defined in terms of
relations provided as background knowledge.

\subsection{First-Order Features as Inputs}

In this paper, we do not want to pre-select input features using an ILP engine.
Instead, we would (ideally) like the inputs to consist of all possible
relational features, and let the network's training process decide on the
features that are actually useful (in the usual manner: a feature that has
0 weights for all out-going edges is not useful).
The difficulty with this is that the number of such features in
first-order logic can be very large, often impractical to enumerate
completely. We clarify first what we mean by a relational feature.

\begin{mydefinition}
{\bf Relational Examples for Classification.}
The set of examples provided can be defined as a binary relation $\mathit{Class}$ which is
a subset of the Cartesian product ${\cal X}\times {\cal Y}$ where ${\cal
X}$ denotes the set of relational instances (for simplicity, and without
loss of generality, we will take each instance to be a ground
first-order term) and ${\cal Y}$ denotes the finite set of class labels. Any single element of the
set of relational examples will be denoted by $Class(a,c)$
where $a \in {\cal X}$ and $c \in {\cal Y}$.
\end{mydefinition}

\begin{mydefinition}
{\bf Relational Features.}
\label{def:rel_feats}
A relational feature is a unary function $F_i : {\cal X} \mapsto
\{\mathit{TRUE}, \mathit{FALSE}\}$ defined 
in terms of a
conjunction of predicates ${Cp}_i(x)$. Each predicate
in ${Cp}_i(x)$ is defined as part of
domain- or background-knowledge $B$.
That is, $F_i(x) \mapsto \mathit{TRUE}$ iff ${Cp}_i(x) \mapsto
\mathit{TRUE}$ and $\mathit{FALSE}$ otherwise.
We will represent each relational feature as a single definite clause $\forall x \; (F_i(x)
\leftarrow {Cp}_i(x))$ in a logic
program, relying on the closed-world
assumption for the complete definition of $F_i$.
We will sometimes call the relational feature $F_i$
simply the \emph{feature} $F_i$, and the definite-clause definition for $F_i$ the
\emph{feature-definition} for $F_i$. If the feature-definition  of $F_i$ is in $B$,
we will sometimes say ``feature $F_i$ is in $B$.'' We will usually
denote the set of features in $B$ as ${\cal F}_B$ or simply ${\cal F}$.
\end{mydefinition}

\begin{mydefinition}
{\bf Classification clause.}
A clause for classifying a relational example
$Class(a,c)$
is a clause $\forall x (Class(x,c) \leftarrow {Cp}(x))$,
where ${Cp}(x)$ is a conjunction of predicates. Each predicate in ${Cp}(x)$ is defined as part of
domain- or background-knowledge $B$. 
\end{mydefinition}

\noindent
It is evident from the definitions that some or
all of a classification
clause can be converted into relational features.

\begin{figure}
\centerline{\includegraphics[width=1\textwidth,height=0.32\textheight]{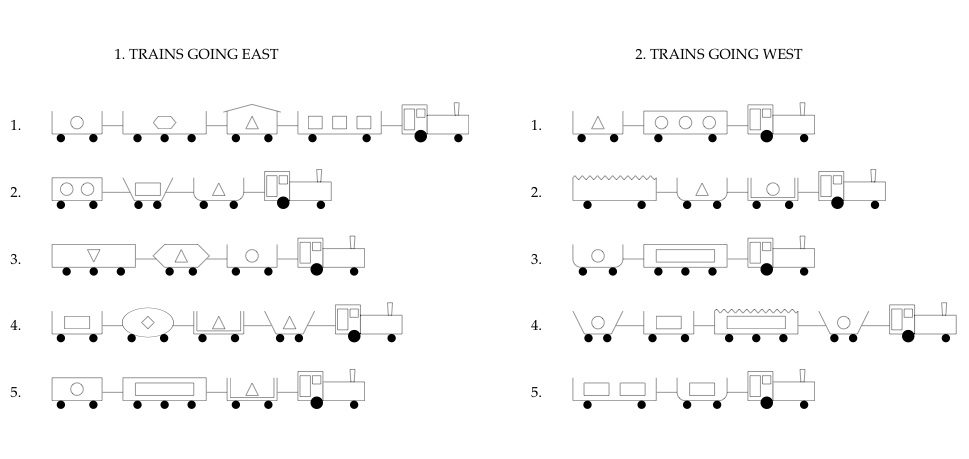}}
\caption{Michalski's trains problem.
There are two sets of trains: Eastbound and Westbound.
Descriptors for the trains include: number, shape and lengths of the car,
shape of load, and so on. The task is to determine classification
rules to distinguish Eastbound trains from Westbound ones. }
\label{fig:trains}
\end{figure}

\begin{myexample}
{\bf The trains problem.}
A well-known synthetic problem in the ILP literature (Michalski's ``Trains'' problem, originally posed in~\cite{michal:trains} -- see Fig.~\ref{fig:trains}) can be used to illustrate this
For this problem each relational example is a pair, consisting of
a ground first-order term and a class label. 
The ground term represents the train (for example 
$Train(Car(Long,$ $Open,Rect,3),$ $Car(Short,Closed,Triangle,1),$ $\ldots)$) and the label denotes whether or not it is Eastbound.

We assume also that we have access to domain (or background) knowledge
$B$ that allows us to access aspects of the relational instance
such as $Has\_Car(Train(\ldots),$ $Car(\ldots))$, $Closed(Car(\ldots))$
and so on. Then a clause for classifying Eastbound trains is (we leave out the
quantifiers for simplicity): 
\[
Class(x,East) \leftarrow Has\_Car(x,y), Short(y), Closed(y)
\]

Here ${Cp}(x)$ = $Has\_Car(x,y) \wedge Short(y) \wedge Closed(y)$. The following relational features  can be
obtained from this classification clause:
\[
F_1(x) \leftarrow Has\_Car(x,y), Short(y)
\]
and:
\[
F_2(x) \leftarrow Has\_Car(x,y), Closed(y)
\]

The classification clause could now be written as:
\[
Class(x,c) \leftarrow F_1(x), F_2(x)
\]
\end{myexample}

\begin{mydefinition}
{\bf Feature Vector of a Relational Instance.}
For a set of features ${\cal F}$ ordered in some canonical sequence
$(F_1,F_2,\ldots,F_d)$ we obtain a Boolean-vector
representation $a' \in \{0,1\}^d$ of
a relational instance $a \in {\cal X}$ using the function $FV: {\cal X} \rightarrow
\{0,1\}^d$, where the
$i^{\mathrm th}$ component ${FV}_{i}(x)$ = 1 if $F_i(x)\mapsto
\mathit{TRUE}$, and 0
otherwise. We will sometimes say $a'$ is
the {\em feature-space\/} representation of $a$.
\end{mydefinition}
 


\subsection{Logical Relevance and Redundancy of Features}

So far, features have been defined purely syntactically.
This is usually not sufficient to ensure
that a feature in the class is relevant to the problem considered, or that it is redundant given another feature.
For example, the feature $F_1$ defined by the clause $F_1(x) \leftarrow Has\_Car(x,y), Open(y), Closed(y))$
is clearly irrelevant in the trains problem, since no car can be both open and closed; and the feature $F_3$ defined by the clause $F_3(x) \leftarrow Has\_Car(x,y),$ $Short(y), Closed(y)$
is redundant given a feature-definition $F_2(x) \leftarrow~Has\_Car(x,y),$ $Closed(y), Short(y)$.

In their full scope, both irrelevance and redundancy of features are semantic notions  dependent on domain knowledge. Since features are defined by clauses, in this paper we will use approximations based on the well-understood concept of clause subsumption from the ILP literature, the main definitions of which we reproduce here for completeness:

\begin{mydefinition}
{\bf Clause subsumption.} We use Plotkin's partial ordering on the set of clauses \cite{plotkin:mi5}.
A clause $C$ subsumes a clause $D$ or $C \preceq_\theta D$, iff there exists a substitution
    $\theta$ s.t. $C \theta \subseteq D$.
It is known that if $C \preceq_\theta D$ then $C \models D$. Further if
$C \preceq_\theta D$ and $D \preceq_\theta C$ we will say $C \equiv_{\theta} D$.
\end{mydefinition}

\begin{myremark}
{\bf Most specific clause in a depth-bounded mode language\/.}
This notion is due to Muggleton \cite{mugg:progol}.
Given a set of modes $M$, let ${\cal L}_d(M)$ be the $d$-depth mode
language. Given background knowledge $B$
and an instance $e$, let the most specific clause
that logically entails $e$ in the sense described in \cite{mugg:progol}
be $\bot(B,e)$. For every $\bot(B,e)$ it is shown in \cite{mugg:progol} that: 
(a) there is a $\bot_d(B,e)$ in ${\cal L}_d(M)$ s.t.
    $\bot_d(B,e) \preceq \bot(B,e)$; and
(b) $\bot_d(B,e)$ is finite.
\end{myremark}

\noindent
Using these notions, we adopt the following definitions:

\begin{mydefinition}
{\bf Logical relevance.} Given a set of examples $E$, and background knowledge $B$,
        an independent clause $C$ is \emph{relevant} if $C \preceq \bot_d(B,e)$ 
        for at least one $e \in E$.
        We will further restrict this to $C \subseteq \bot_d(B,e)$.
\end{mydefinition}

\begin{mydefinition}
{\bf Logical redundancy.} For a pair of independent clauses $C$ and $D$, we will
        say $D$ is \emph{redundant} given $C$ (and {\em vice-versa\/}), 
        if $|C| = |D|$ and $C \equiv_{\theta} D$. 
\end{mydefinition}

\noindent
These definitions based on subsumption are an
approximation to the true logical definitions of relevance and redundancy, and
can result in some errors (assuming relevance when it does not exist, and missing
redundancy when it does exist). Nevertheless, they can be computed reasonably efficiently.
They form the basis of a method of randomised construction of
inputs for the deep network.

\subsection{Randomised Feature Selection}
\label{sec:sample}

\begin{figure}
\centerline{\includegraphics[height=0.36\textheight,width=0.8\textwidth]{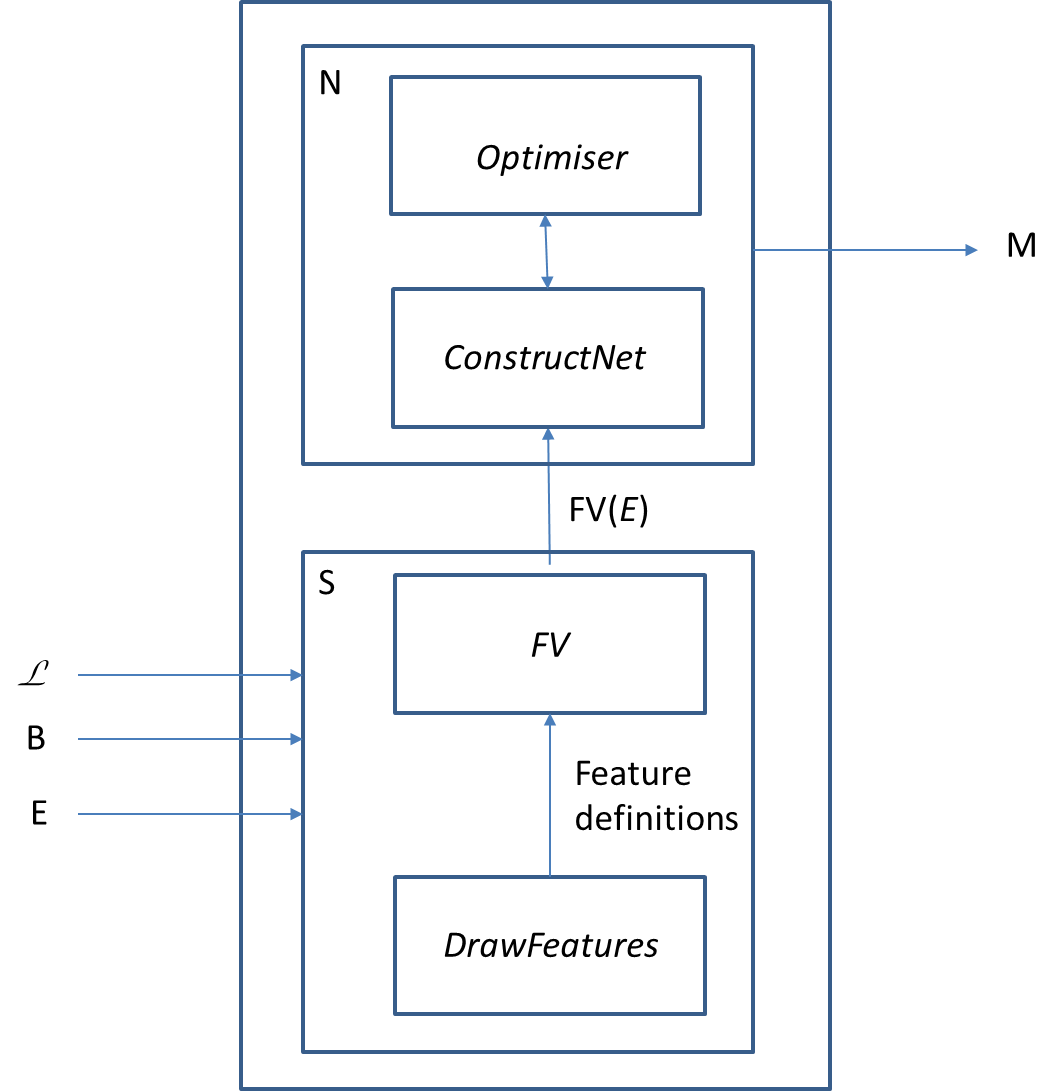}}
\caption{More details on the training component in Fig.~\ref{fig:bigpic}. 
For simplicity, we are taking the modes ${\cal M}$ to be part of the language specification ${\cal L}$, and do not show some parameters. The optimiser finds the best network structure and parameters using the feature-vectors provided for training. FV($E$) denotes the feature-vectors of the relational instances $E$. The model $M$ is a DRM.}
\label{fig:smallpic1}
\end{figure}

The mechanisms for detecting
logical relevance and redundancy help to reduce the number of possible features within
a mode language. But they still do not guarantee that the numbers will
be manageable. We therefore resort to a rejection-based sampling strategy for selecting
inputs (Fig. \ref{fig:feat}). Of course, inputs to the DRM are not the features themselves
but the feature-space representations for the relational instances in $E$ (Fig.~\ref{fig:smallpic1}).

\begin{figure*}[htb]
{\small{
$DrawFeatures(B,E,{\cal M},{\cal L},d,MaxDraws):$
    \begin{description}
        \item[Given:] Background knowledge $B$;
                examples $E$;
                a set of modes ${\cal M}$;
                language constraints ${\cal L}$;
                a depth-bound $d$; and
                an upper-bound on the number of samples $MaxDraws$.
        \item[Find:] A set of features ${\cal F}$ s.t. $|{\cal F}| \leq MaxDraws$                 
    \end{description}
\begin{enumerate}
\item Let $F = \emptyset$
\item Let $draws = 0$
\item Let $i=1$
\item Let $Drawn$ be $\emptyset$
\item {\bf while} $draws \leq MaxDraws$ {\bf do}
\begin{enumerate}
\item Randomly draw with replacement an example $e_i \in E$.
	Let $e_i = Class(a,c)$, where $a$ is a relational instance
    and $c$ is a class label.
\item Let $\bot_d(B,e_i)$ be the most specific clause in
    the depth-limited mode language ${\cal L}_d({\cal M})$ that subsumes $\bot(B,e_i)$ (i.e., the
    most-specific clause that entails $e_i$, given $B$).
\item Randomly draw a clause $C_i$ s.t. $C_i \subseteq \bot_d(B,e_i)$ \label{step:draw}
\item {\bf if} ($C_i$ is not redundant given $Drawn$) {\bf then}
    \begin{enumerate}
        \item Let $C_i = (Class(x,c) \leftarrow {Cp}_i(x))$
        \item Let $F_i = (F_i(x) \leftarrow {Cp}_i(x))$
        \item $F~:=~ F \cup \{F_i\}$
        \item $Drawn ~:=~ Drawn \cup \{C_i\}$
        \item increment $i$
    \end{enumerate}
\item increment $draws$
\end{enumerate}
\item {\bf done}
\item return $F$
\end{enumerate}
}}
\caption{A randomised procedure for drawing features from a depth-limited mode language.
    A procedure for randomly drawing clauses subsuming a most-specific clause
    required in Step \ref{step:draw} is described in \cite{filip:mcmlj}.}
\label{fig:feat}
\end{figure*}

\section{Logical Explanations for Predictions by a DRM}
\label{sec:expl}

\begin{figure}
\begin{center}
\includegraphics[height=0.36\textheight,width=\textwidth]{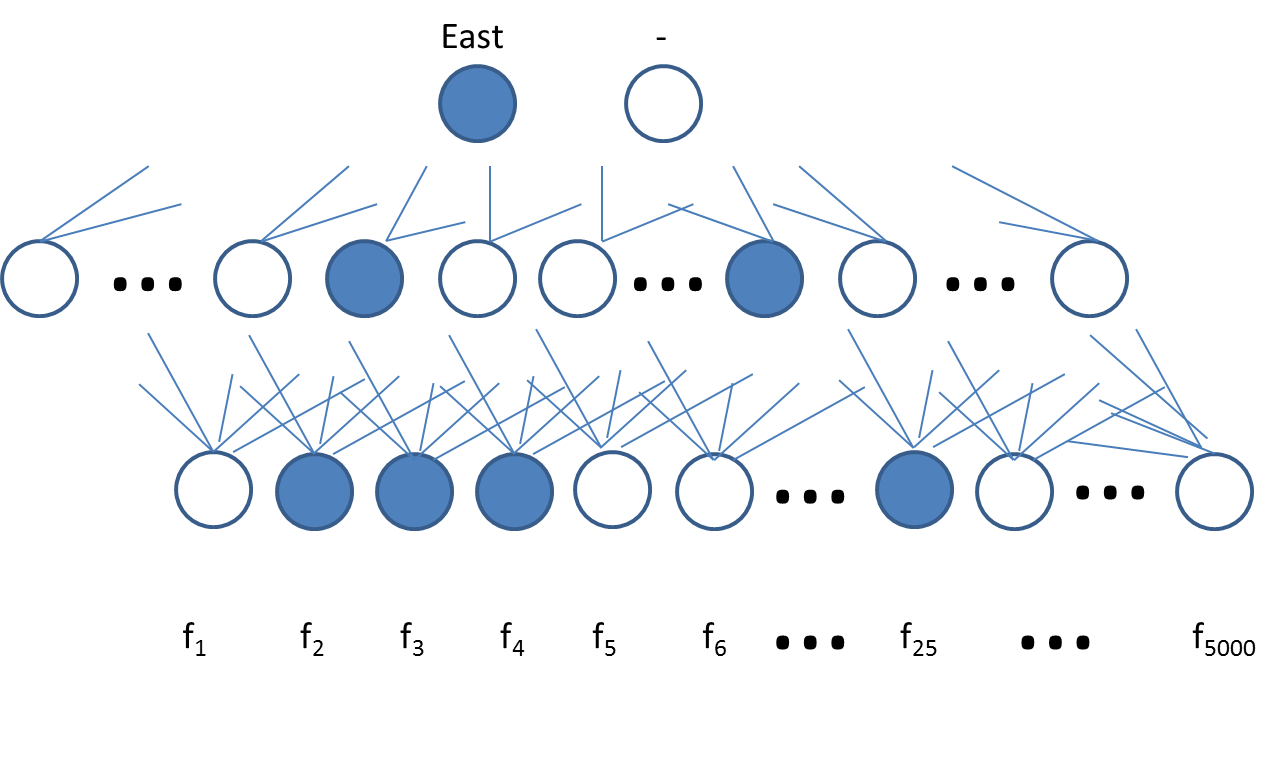}
\end{center}
\caption{Prediction of the class of a relational instance $a$ by a DRM. The inputs $f_i$ are the component values of the feature-vector ${FV}_i(a)$. The shaded inputs
denote a feature-value of 1. With a DRM, there is more
to $f_i = 1$ than just an assignment of $1$ to feature:
it also means that some conjunction of background
predicates ${Cp}_i(a)$ is {\em TRUE\/} for the relational
instance $a$. The shaded circles in hidden
layers denote activated nodes. The shaded output node denotes
the prediction of the relational instance is $East$.}
\label{fig:drmpred}
\end{figure}

Predicting the class-label of a relational instance $a$ 
using the model $M$ in Fig.~\ref{fig:smallpic1} is a 2-step
process: (1) The feature-vector representation $a'$ of $a$
is obtained using the feature-definitions found during the
training stage; and (2) $a'$ is provided as input to the
model $M$ found during training, which then computes the
class-label (Fig.~\ref{fig:drmpred}).

It has long been understood that neural models can compute accurate answers for questions like ``What is the prediction for {\bf x}?''. But what can
be said about why the answer is what it is? Unsurprisingly,
there has been a lot of research effort invested into extracting explanations for the predictions made by
a neural network (see \cite{gabbay:neurosym}, Chapter 3
for a full description of this line of research). 
Most of this effort has been in the direction of translating the network into
a single symbolic model that guarantees correspondence to the neural model (for example, \cite{thrun:neuralcompre}).
Alternatively, a single symbolic model can be learned
that approximates the behaviour of the neural model over all inputs (for example, \cite{craven:neuralcompre}). Although
distinct in their aims, both approaches still
result in two separate models, one neural and the
other symbolic. In both cases the symbolic model
is intended to be a readable proxy for the neural model, which can then
form a basis for an explanation of ``why'' questions. 
But there are some inherent trade-offs:

\begin{itemize}
	\item If we try to replicate exactly the behaviour of the
    	neural network with a symbolic model (as is done,
        say, in \cite{thrun:neuralcompre}), then the resulting
        model may not be any more comprehensible than the
        neural network; and
 	\item If we try only to approximate the behaviour of
    	the network using logical predicates
        (as is done, say, in \cite{craven:neuralcompre}),
        then we run the risk of not being able to
        replicate the network's behaviour sufficiently
        accurately over all instances, because
        of inadequacies of the logical predicates available.
\end{itemize}

Both these issues are exacerbated for modern-day deep networks,
with many hidden layers and large numbers of inputs. One
way to side-step these difficulties is simply to drop
the requirement of translating the \emph{entire} network model into a single symbolic model.  Recent research called LIME
(``Local Interpretable Model-agnostic Explanations'': \cite{lime})
proposes producing readable proxies ``on-demand'',
for any kind of black-box predictor. The key feature is that a LIME-style explanatory model is constructed
only when a prediction for an instance is sought, and the explanation is required to be faithful to the black-box's
(in our case, a neural network) predictions for the instance
and its near-neighbours. The intuition is that while comprehensible models may not be possible for the entire black-box model, they may be possible when restricted to local predictions.

The requirement of having to be consistent with only local
instances results in a different kind of problem. Since the
near-neighbours of an instance may be quite small,
there may be insufficient constraints, in data-theoretic
terms, to narrow down on a unique (or even a small
number of) explanations. So, how then is an explanation
to be selected? In LIME, this is left to the
loss-function. In Bayesian terms, this means defining
an appropriate prior to guide selection when the data are 
insufficient.

\subsection{Local Explanations}

We now consider a logical formulation
of the setting introduced in the LIME paper that is well-suited
to explaining DRM models.
We draw on specific
situations described in \cite{mugg:sigart}, namely
to clarify what is meant by a
local explanation, and its data-theoretic evaluation.
Later we will introduce a
domain-dependent prior to develop a Bayes-like
selection of interpretable explanations.

\begin{mydefinition}
{\bf Explanations.}
In this paper, we will only be concerned with
explanations for the classification of an instance.
As before, we assume a set of relational instances
${\cal X}$, and a finite set of classes ${\cal Y}$.
Given a relational instance $a \in {\cal X}$ and
a label $c \in {\cal Y}$,
the statement $Class(a,c)$ will denote that the class of $a$ is $c$.
Let $e = Class(a,c)$. Then given background knowledge $B$, an explanation for $e$ is
a set of clauses $H$ s.t. $B \cup H \models e$.
\end{mydefinition}

\noindent
At this point, we differ from Good \cite{good:expl}, for whom
it is neither necessary nor sufficient for good explanations to logically imply $e$, given $B$.
As we present it here, logical implication is necessary, but as will be seen below, not sufficient for a good
explanation.
We will seek explanations of a restricted kind, namely
those defined in terms of features. 

\begin{mydefinition}{\bf Feature-clauses.}
Let ${\cal F}$ be a set of features.
A definite clause is said to be a feature-clause
if all negative literals are of the form $F_i(x)$, where the $F_i$ are in ${\cal F}$ and the $x$ is universally quantified.
\end{mydefinition}

\subsubsection{Single-Clause Explanations}
\label{sec:singexp}

Assume we have a set of $d$ features ${\cal F}$ with
corresponding feature-definitions in the background
knowledge $B$. We will first consider the case where
explanation $H$ consists of a single feature-clause.
That is, $H = \{C\}$, where $C$ is a definite
clause $\forall x \; (Class(x,c) \leftarrow Body)$.
Here $Body$ is a conjunction of ${F}_i/1$ literals,
each of which is defined in $B$.
The ILP practitioner will recognise that finding
a single feature-clause explanation for $Class(a,c)$ is an
instance of the
``Single example clause, single hypothesis clause''
situation identified in \cite{mugg:sigart},
which forms the basis of explanation-based learning (EBL)~\cite{ebg:mitchell}.

\begin{myexample}
{\bf An explanation in the trains problem.}
\label{ex:expl_trains}
An explanation $H = \{C\}$ for the first train in the left column of Fig.~\ref{fig:trains} is the feature-clause:

\[
C: Class(t1,East) \leftarrow {F}_1(t1), {F}_2(t1), {F}_3(t1)
\]
\noindent
where $B$ contains:
\[
\forall x ({F}_1(x) \leftarrow \exists y (Has\_Car(x,y), Short(y)))
\]
\[
\forall x ({F}_2(x) \leftarrow \exists y (Has\_Car(x,y), Closed(y)))
\]
\[
\forall x ({F}_3(x) \leftarrow \exists y (Has\_Car(x,y), Long(y)))
\]

\noindent
Here $t1$ is used as short-form for a structured term, describing the train, and,
as before, let us assume that appropriate definitions exist in $B$ for predicates like
$Has\_Car$, $Long$, $Short$, and $Closed$ to succeed (or fail) on terms like $t1$.

\end{myexample}

One explanation for $e = Class(a,c)$ can be obtained immediately as follows.
Consider the set
$F_a = \{F_i: F_i \in {\cal F} ~\mathrm{s.t.}~{FV}_i(a) = 1\}$
(let us call these the set of {\em active\/} features
for $a$). By definition, if ${FV}_i(a) = 1$ then
$F_i(a) \mapsto \mathit{TRUE}$. For simplicity, let us
suppose that $F_a$ = $\{F_1,\ldots F_k\}$ is the set of
active features for $a$,
and let $C:~Class(a,c) \leftarrow {F}_1(a), {F}_2(a),\ldots,{F}_k(a)$.
Assuming the $F_i$ are defined in $B$,
$H ~=~ \{C\}$ is an explanation for $e$.\footnote{ILP practitioners will recognise $C$ as
being analogous to the most-specific clause in \cite{mugg:progol},
and we will call it the most-specific feature-clause for $e$, given ${\cal F}$ and $B$.}
Fig.~\ref{fig:unstruct} is a simple procedure to construct a single-clause
explanation $C'$ that relies on the result that if
$C' \preceq_\theta C$, then $C' \models C$. Therefore, if $B \cup \{C\} \models e$,
then evidently $B \cup \{C'\} \models e$. That is, $\{C'\}$ will also be an
explanation for $e$. For reasons that will be apparent, we will call $\{C'\}$ an
``unstructured'' explanation.\footnote{
In practice, the lattice in Step \ref{step:lattice} would be represented by a graph,
and finding an element of the lattice in Step \ref{step:search} will involve some
form of optimal graph-search to find an optimal (or near-optimal) solution.More on this
later.}

\begin{figure*}[htb]
{\small{
$ConstructUnstruct(e,B,{\cal F}):$
    \begin{description}
    \item[Given:] A relational instance $e$;
        background knowledge $B$;
        a set of features ${\cal F}$ with definitions in $B$.
    \item[Find:] A single feature-clause explanation $H$ s.t. $B \cup H \models e$.
    \end{description}
\begin{enumerate}
	\item Let $e = Class(a,c)$
    \item Let $a' = FV(a)$
     \item Let ${\cal F}'$ be the set of features
        that map to $TRUE$ in $a'$ \label{step:bot}
    \item Let ${\cal H}$ be the subset-lattice of ${\cal F}'$ \label{step:lattice}
    \item Let $F'$ be any element in ${\cal H}$ s.t. \label{step:search}
            \begin{itemize}
                \item $Body$ is the conjunction of features in $F'$;
                \item $C' = \forall x(Class(x,c) \leftarrow Body)$
            \end{itemize}
    \item return $\{C\}$
\end{enumerate}
}}
\caption{A procedure for identifying a single-clause unstructured explanation.}
\label{fig:unstruct}
\end{figure*}

\subsubsection{Multi-Clause Explanations}
\label{sec:multexp}

We now extend explanations to a restricted form of multi-clause
explanations, which use features that are not all defined in
the background knowledge $B$.
We will require these ``invented'' features to be (re-)expressible
in terms of features already defined in $B$. In this paper
we will require structured and unstructured explanations to
be related by the
logic-program transformation operations of \emph{folding} and {\emph{unfolding}.
That is, given an unstructured explanation $H$ and a  structured explanation
$H_1$, $H$ can be derived from $H_1$ using one or more unfolding transformations;
and $H_1$ can be derived from $H$ using one or more folding transformations.
We describe the transformations, along with the conditions that ensure that
the computable answers do not change when the transformation is applied.



\begin{mydefinition} \label{def:unfold}
{\bf (One-Step) Unfolding of a clause \cite{hogger:lpbook}.}
Given a set of definite clauses $P$, W.l.o.g. let $C \in P$ s.t.
$C$ = $Head \leftarrow L_1,\ldots,L_i,\ldots,L_k$ ($k \geq 1$). Let
$C'$ be a clause ${L'}_i \leftarrow {Body}'$ s.t.
$L_i$ and ${L'}_i$ unify with m.g.u $\theta$. Then 
the (one-step) unfolding of $C$ w.r.t $L$ using $P$ is the clause
$Unfold(C) : (Head \leftarrow L_1,\ldots,L_{i-1},{Body}',L_{i+1},\ldots,L_k)\theta$
\end{mydefinition}

\begin{mydefinition} \label{def:fold}
{\bf (One-Step) folding of a clause \cite{hogger:lpbook}.}
Given a set of definite clauses $P$, W.l.o.g. let $C \in P$ s.t.
$C$ = $Head \leftarrow L_1,\ldots,{L_{i-1},Body}_i,L_{i+1},\ldots,L_k$, where
${Body}_i$ is some literals $B_1,B_2,\ldots,B_j$ ($j,k \geq 1$). Let
$C'$ be a clause ${L'}_i \leftarrow {Body}'$ s.t. there is a substitution
$\theta$ that satisfies:
(a) ${Body}_i ={Body}'_i \theta$;
(b) Every existentially quantified variable $y$ in $C$, $y \theta$ is
	a variable that occurs only in ${Body}_i$ and nowhere else;
(c) For any pair of distinct existential variables $y,z$ in $C$
	$y\theta \neq z \theta$; and
(d) $C'$ is the only clause in $P$ whose positive literal unifies with ${L'}_i \theta$.
Then the (one-step) unfolding of $C$ w.r.t. ${Body}_i$ using $P$ is 
the clause
$Fold(C): (Head \leftarrow L_1,\ldots,L_{i-1},{L'}_i\theta,L_{i+1},\ldots, L_k)$.
\end{mydefinition}

\begin{myremark}
{\bf Correctness of Transformations.}\label{rem:trans_correct}
In \cite{gabbay:lpbook} the unfold and fold transformations are defined above
are shown to be correct as replacement rules w.r.t. the minimal-model (MM) semantics.
That is, if $P$ is a definite-clause program and $C \in P$,
then 
$MM(P)$ =
$MM((P - \{C\}) \cup Trans(C))$, where
$Trans(C)$ is $Fold(C)$
or $Unfold(C)$.
\end{myremark}

\noindent
We use this to construct structured explanations that are
correct, in the computational sense just identified. That is,
a structured explanation can replace an unstructured explanation
without altering the minimal model of the (program containing) the
unstructured explanation. It is convenient for us to introduce
the notion of an ``invented'' feature.

\begin{mydefinition} \label{def:inv_feat}
{\bf Invented Feature.}
Given background knowledge $B$, a feature $F$ is said
to be an invented feature if: (1) $F$ is not defined in $B$;
and (2) The definition of $F$ is a feature-clause $C$ whose body
contains features in $B$ only; or is a clause that
unfolds to a feature-clause $C$ containing features only in $B$.
We will sometimes denote $C$ as $UFC(F)$ (short for
``unfolded feature-clause for F'').
\end{mydefinition}

\begin{myexample}
{\bf Invented features in the trains problem.}
The features $F_{1,1}$ and $F_{1,2}$ are invented features:

\begin{tabular}{l} \\
${F}_{1,1}(x) \leftarrow {F}_{2}(x), {F}_{1,2}(x)$  \\
${F}_{1,2}(x) \leftarrow {F}_4(x), {F}_9(x)$\\
\end{tabular}

\noindent
where the $F_i$ are features defined in background knowledge:

\begin{tabular}{l} \\
${F}_2(x) \leftarrow Has\_Car(x,y), Short(y)$\\
${F}_4(x) \leftarrow Has\_Car(x,y), Wheels(y,3)$ \\
${F}_9(x) \leftarrow Has\_Car(x,y), Load(y,triangle)$ \\
\end{tabular}

\noindent
Then $UFC(F_{1,1})$ is:
\[
	F_{1,1,}(x) \leftarrow F_2(x), F_4(x), F_9(x)
\]
\noindent
and $UFC(F_{1,2})$ is identical to the definition of $F_{1,2}$.
\end{myexample}

\noindent
We distinguish between unstructured and structured definitions based
on the presence or absence of invented features.

\begin{mydefinition}{\bf Unstructured and structured explanations.}
\label{def:struct}
Let $a \in {\cal X}$ and ${\cal Y}$ be a set of class labels,
with $c \in {\cal Y}$.
Let $N:\Re^d \rightarrow {\cal Y}$ be a
predictive model such that $N(FV(a)) = c$.
Given background knowledge $B$,
let $H$ be an explanation for $Class(a,c)$ containing
the feature-clause $C:\forall x(Class(x,c) \leftarrow Body)$.
Let $Body$ consist of features ${\cal F}_C$.
Let ${\cal F}_C$ be partitioned into:
(a) ${\cal F}_{C,old}$, consisting of features defined in $B$; and
(b) ${\cal F}_{C,new}$, consisting of features defined in $H - B$.

We will call $H$ an \emph{unstructured explanation} iff:
(1) ${\cal F}_C$ contains only old features (that is,
                ${\cal F}_{C,new} = \emptyset$); and
(2) $H$ = \{C\}.

We will call $H$ a \emph{structured explanation} iff:
(1) ${\cal F}_C$ contains only invented features (that is,
        ${\cal F}_{C,old} =  \emptyset$); and
(2) $H = \{C\} \cup InvF$, where $InvF$ pnly contains clauses
    defining invented features in ${\cal F}_{C,new}$;
(3) Each feature $F$ in ${\cal F}_{C,new}$ there is a single
	clause definition in $InvF$, s.t. $F$ unfolds to a
    unique feature-clause defined using features in $B$ only; and
(4) At least one $F \in {\cal F}_{C,new}$ unfolds
    to a feature-clause that contains at least 2 features
    from $B$ (that is, there is at least one invented feature
    that is not a trivial rewrite of features in $B$).\footnote{
    We assume the case where ${\cal F}_{C,old} \neq \emptyset$ and
    ${\cal F}_{C,new} \neq \emptyset$
will be represented by a structured explanation in which there is a new feature
representing the conjunction of existing features. See the example that follows.}
\end{mydefinition}

That is, a structured explanation is a set of clauses containing a classification clause
$C$ along with definitions of invented features (and thus is a case
of the ``single example clause, multiple hypothesis clauses'' situation in \cite{mugg:sigart}).

\begin{myexample}
{\bf Unstructured and structured explanations for the trains problem.}\label{ex:strucexptrains}
Suppose we are given an instance $a \in {\cal X}$,
and a set of features ${\cal F}$,
each defined by feature-definitions
in background $B$. Then we will call the following
an unstructured explanation ($x$ is universally quantified):
\[
C: Class(x,East) \leftarrow {F}_2(x), {F}_{3}(x), {F}_{4}(x), {F}_{9}(x)
\]

\noindent
The following is a structured explanation $H_1$:

{\small{
\begin{tabular}{l} \\
$Class(x,East) \leftarrow {F}_{1,1}(x), {F}_{1,2}(x)$\\
${F}_{1,1}(x) \leftarrow {F}_{2}(x), {F}_{3}(x)$  \\
${F}_{1,2}(x) \leftarrow {F}_4(x), {F}_9(x)$\\
\end{tabular}
}}

\vspace*{0.2cm}
\noindent
Note that $H_1$ unfolds to $H$.
Another structured explanation that also unfolds to $H$ is $H_2$ below:

{\small{
\begin{tabular}{l} \\
$Class(x,East) \leftarrow {F}_{1,1}(x), {F}_{1,2}(x)$\\
${F}_{1,1}(x) \leftarrow {F}_{1}(x), {F}_{4}(x)$  \\
${F}_{1,2}(x) \leftarrow {F}_3(x), {F}_9(x)$\\
\end{tabular}
}}

\vspace*{0.2cm}
\noindent
We will assume that explanations of the form
$Class(x,East) \leftarrow {F}_{1,1}(x), {F}_3(x), {F}_9(x)$
will be represented as $H_2$. Also, by definition, the following
is not a structured explanation, since all new features are trivial
rewrites of existing ones:

{\small{
\begin{tabular}{l} \\
$Class(x,East) \leftarrow {F}_{1,1}(x), {F}_{1,2}(x)$\\
${F}_{1,1}(x) \leftarrow {F}_{1}(x)$  \\
${F}_{1,2}(x) \leftarrow {F}_3(x)$\\
\end{tabular}
}}

\vspace*{0.2cm}
\noindent
But this is a structured explanation:

{\small{
\begin{tabular}{l} \\
$Class(x,East) \leftarrow {F}_{1,1}(x), {F}_{1,2}(x)$\\
${F}_{1,1}(x) \leftarrow {F}_{1}(x)$\\
${F}_{1,2}(x) \leftarrow {F}_3(x), {F}_9(x)$\\
\end{tabular}
}}
\end{myexample}

\noindent
It is obvious enough that the unfolding step that every
structured explanation allows the derivation
of a correct unstructured explanation.

\begin{myremark}
{\bf Deriving an Unstructured Explanation from a Structured Explanation.} \label{rem:unstruc_struc}
Let $H = \{C'\} \cup InvF$ be a structured explanation for a
relational example $Class(a,c)$ given background knowledge $B$. W.l.o.g. from Defn.~\ref{def:struct}, 
$C'$ is a clause $Class(x,c) \leftarrow \{{F'}_1(x), {F'}_2(x),\ldots,{F'}_k(x)$
s.t. each ${F'}_i$ unfolds using $InvF$ to a unique feature-clause
${F'}_i(x) \leftarrow F_{i,1}(x), \ldots, F_{i,n_i}(x)$, where the $F_{i,\cdot}$ are
defined in $B$. From Defn.\ref{def:unfold}, unfolding $C'$ w.r.t. the ${F'}_i$ using
$InvF$ results in the clause $C:~Class(x,c) \leftarrow F_{1,1},\ldots, F_{k,n_k}$.
From Remark \ref{rem:trans_correct}, the minimal model of $B \cup H$ will be unchanged
by replacing $C'$ with $C$. It follows that $H$ = $\{C\}$ is an  unstructured explanation for $Class(a,c)$ that is computationally equivalent to $H'$.
\end{myremark}

\noindent
One or more correct structured explanations follow from
an unstructured if the conditions in Defn.\ref{def:fold}
hold. Constraints on the invented features can ensure
this.

\begin{myremark}
{\bf Deriving Structured Explanations from an Unstructured Explanation.}
\label{rem:struc_unstruc}
Let $H = \{C\}$ be an unstructured explanation for a
relational example $Class(a,c)$ given background knowledge $B$. W.l.o.g. from Defn.~\ref{def:struct}, 
$C$ is a clause $Class(x,c) \leftarrow \{{F}_1(x), {F}_2(x),\ldots,{F}_n(x)$
s.t. each ${F}_i$ is defined in $B$. Let $InvF$ 
consist of $k$ features ${F'}_1,\ldots,{F'}_k$ s.t.
each ${F'}_i$ is uniquely defined by features in a block of
a $k$-partition of $F_1,\ldots,F_n$. Let $C'$ be
the clause $Class(x,c) \leftarrow {F'}_1(x), {F'}_2(x),\ldots,{F'}_k(x)$.
Then $H' = \{C'\} \cup InvF$ is a structured explanation for $Class(a,c)$
that is computationally equivalent to $B \cup \{C\} \cup InvF$.
This follows since the conditions (a)--(d) in Defn. \ref{def:fold} are trivially
satisfied (condition (a) follows with $\theta$ being a simple renaming substitution;
(b)--(c) follow since there are no existential variables in the definitions of the
${F'}_i$; and
(d) since there is a single clause definition for each feature in $InvF$).
\end{myremark}

Remark \ref{rem:struc_unstruc} suggests a straightforward non-deterministic procedure
to construct a correct
structured explanation (Fig.~\ref{fig:struct}).

\begin{figure*}[htb]
{\small{
$ConstructStruct(H,B,k):$
    \begin{description}
        \item[Given:] An unstructured explanation $H$; background knowledge $B$; and a number $k$ ($\geq 2$)
        \item[Find:] A structured explanation $H'$ that is computationally
        	equivalent to $H$
    \end{description}
\begin{enumerate}
\item Let $H$ = $\forall x(Class(x,c) \leftarrow Body)$ 
\item Let $F$ be the set of features in $Body$
\item Let $P_k$ be a set s.t:
	\begin{enumerate}
		\item[(a)] \label{step:norewrite} Either $P_k$ is a $k$-partition of $F$ s.t.
        	at least one block in $P_k$
        	contains $2$ or more elements; or
  		\item[(b)] Or, $P_k = \emptyset$ (if no such $k$-partition exists)
 	\end{enumerate}
\item If $P_k = \emptyset$ return $\emptyset$.
\item Otherwise:
\begin{enumerate}
    \item Let $P_k$ consist of blocks $b_1,b_2,\ldots,b_k$
    \item For each block $b_i \in P_k$:
    \begin{enumerate}
        \item Construct a new feature ${F'}_i$ with
            definition $C_i$ = $\forall x({F'}_i(x) \leftarrow {Body}_i)$,
            where ${Body}_i$ is the conjunction of the features in $b_i$
    \end{enumerate}
    \item Let $C_0$ = $\forall x(Class(x,c) \leftarrow {Body}_0)$ where
            ${Body}_0$ is the conjunction of the ${F'}_1,{F'}_2,\ldots,{F'}_k$
    \item Let $H'$ = $\bigcup_{i=0}^k C_k$
\end{enumerate}
\item Return $H'$.
\end{enumerate}
}}
\caption{A procedure for obtaining a structured explanation from an unstructured
	explanation, by inventing $k$ features. The condition in Step \ref{step:norewrite} is
    not required by Remark \ref{rem:struc_unstruc}, but prevents inventing
    features that are trivial rewrites of existing features (see Defn. \ref{def:struct}).}
\label{fig:struct}
\end{figure*}

All explanations so far have only been required to explain a single
relational instance. We extend this to explain the prediction by a black-box
classifier of relational instances
that are ``close'' to each other. This is the requirement identified
in LIME \cite{lime}. 
Readers familiar with the ILP literature will recognise the task of finding
such local explanation as as corresponding to either
``Multiple examples, single hypothesis clause'' or
``Multiple examples, multiple hypothesis clauses'' situations
identified in \cite{mugg:sigart}
(depending on whether unstructured or structured
explanations are constructed).

\subsection{Local Explanations for a Black-Box Predictor}
\label{sec:localexp}

We are specifically interested in constructing explanations for
the classification of an instances resulting from
some (opaque) predictive model.

\begin{mydefinition}
{\bf Explanation for a prediction by a model.}
Given a relational instance $a \in {\cal X}$ and a label $c \in {\cal Y}$.
Assume we have a set of $d$ features ${\cal F}$.
Given a predictive model $N:\Re^d \rightarrow {\cal Y}$,
we will say $H$ is an explanation for $N(FV(a)) = c$ if
$H$ is an explanation for $e = Class(a,c)$.
\end{mydefinition}

\noindent
We have just seen how to construct the most-specific
feature-clause for $e = Class(a,c)$ from the
set of active features for $a$. 
In fact, we will now want a little bit more from an explanation.
Following \cite{lime}, we will seek explanations that are
not only consistent with a predictive model on an instance $x$, but be consistent
with predictions made by the predictive model in a \emph{local neighbourhood} of $x$.

\begin{mydefinition}{\bf Neighbourhood.}
Given relational instance $a \in {\cal X}$, and a set of $d$ features ${\cal F}$,
and given some $\epsilon \in \Re$, we denote the neighbourhood of $a$ as
$Nbd(a) ~=~ \{x: x \in {\cal X}~{\mathrm and}~{Dis}(FV(a),FV(x)) \leq \epsilon\}$.
Here $Dis$ is some appropriate distance measure defined over $d$-dimensional
vectors.
\end{mydefinition}

\noindent
In practice the dimensionality $d$ can be quite large, and since the standard Euclidean distance
is known to be problematic in high-dimensional spaces~\cite{aggarwal:dist} we will use an
alternative measure.

\begin{mydefinition}{\bf Locally Consistent Explanations.}\label{def:local}
Given relational instance $a \in {\cal X}$, and a predictive model $N$, let $N(FV(a)) = c$. We
define the following subsets of $Nbd(a)$:
$E^{+}(a) = \{b: b \in Nbd(a) ~{\mathrm {and}}~ N(FV(b)) = c)\}$
and $E^{-}(a) = \{b: b \in Nbd(a) ~{\mathrm {and}}~ N(FV(b)) \neq c \}$.
Then an explanation $H$ for $Class(a,c)$ given $B$ is a locally consistent explanation if:
(1) for each $a' \in E^{+}(a)$ $H$ is an explanation for $Class(a',c)$
(that is, $B \cup H \models  Class(a',c)$); and
(2) for each $a' \in E^{-}(a)$
$B \wedge H \wedge \neg Class(a',c) \not \models \Box$.
\end{mydefinition}

\begin{myexample}
{\bf Locally consistent explanation for the trains problem.}
For simplicity, we consider the situation where $E^{-} = \emptyset$.
Suppose we now know that the local neighbourhood of the first train ($t1$) in the left column
of Fig.~\ref{fig:trains} only contains the second train ($t2$) in that column. Let us assume that
${\cal F}$ consists just of the functions defined in 
Example~\ref{ex:expl_trains}. With those definitions, and a predictive model $N$,
let $N(FV(t1)) = N(FV(t2))$ ($= East$, say). Then $E^{+}(t1) = \{t1,t2\}$.

The most-specific feature-clause for $Class(t1,East)$ given ${\cal F}$ and $B$ is:
\[
C1: Class(t1,East) \leftarrow {F}_1(t1), {F}_2(t1), {F}_3(t1)
\]
and for $Class(t2,East)$ is:
\[
C2: Class(t2,East) \leftarrow {F}_1(t2), {F}_2(t2)
\]
\noindent
where the function definitions are as before. Then a locally consistent explanation for
$Class(t1,East)$ is the least-general-generalisation (or Lgg) of $C1$ and $C2$:
\[
Lgg(C1,C2): \forall x (Class(x,East) \leftarrow {F}_1(x), {F}_2(x))
\]

\end{myexample}

In general, $E^{-} \neq \emptyset$, and results from ILP tell us that it may not be possible to find
a single clause that is locally consistent. We next describe a simple,
qualitative form of Bayes Rule that combines likelihood of the data,
with a relevance-based prior preference over explanations. 

it is useful before we proceed further
to have a numerical measure of the extent to which an explanation is
locally consistent.

\begin{mydefinition}\label{def:fidelity}
{\bf Fidelity.}
Let $a \in {\cal X}$. Given a predictive model $N$, let $N(FV(a)) = c$
and $E^{+}(a)$, $E^{-}(a)$ as before. Let $D = (E^{+},E^{-})$ and
$H$ be an explanation for $Class(a,c)$ given background knowledge $B$. Let:
(1) $AgreePos(H)$ = $\{b: b \in E^{+}(a)$ and $B \wedge H \models Class(b,c)\}$; and
(2) $AgreeNeg(H)$ = $\{b: b \in E^{-}(a)$ and $B \cup H \wedge \neg Class(b,c) \not \models \Box\}$.
Then $\mathit{Fidelity}(H|D,B) = F_{H,a} = \frac{|AgreePos(H)|+|AgreeNeg(H)|}{|E^{+}(a)| + |E^{-}(a)|}$.
\end{mydefinition}

\noindent
We note that $AgreePos(H)$ and $AgreeNeg(H)$ are the same as true positives and
true negatives in the classification literature, and so \emph{fidelity} is
a localized form of accuracy (in \cite{lime} the term ``local fidelity''
is used for a localized form of error). 
We note also
that if a structured explanation is derived from an unstructured one,
then fidelity will not change.

\begin{myremark}{\bf Structuring preserves fidelity.}
\label{rem:struct_fidelity}
Let $H'= \{C'\} \cup InvF$ be a structured explanation
derived as in Remark \ref{rem:struc_unstruc}
from an unstructured explanation $H = \{C\}$ for a relational
example $Class(a,c)$, given definite-clauses $B$ and a predictor $N$ s.t.
$N(FV(a)) = c$. Let $E^{+}(a)$
and $E^{-}(a)$ denote relational examples obtained from the neighbourhood of $a$ as
before, and $D = (E^{+}(a),E^{-}(a))$. Then, $Fidelity(H'|D,B) = Fidelity(H|D,B)$.\\

Let $S$ be any finite set of ground $Class/2$ facts and $MM(P)$ be the minimal
model of a definite-clause program $P$. Since $InvF$ is a definite-clause containing definitions of invented features,
$S \cap MM(B \cup  \{C\}) = S \cap MM(B \cup \{C\} \cup InvF)$. 
Now, $AgreePos(H)$ = $\{Class(a',c): Class(a',c) \in (E^{+}(a) \cap MM(B \cup H)$
and
$AgreeNeg(H)$ = $\{Class(a',c): Class(a',c) \in (E^{-}(a) - (E^{-}(a) \cap MM(B \cup H)$.
Since $E^{+}(a)$ and $E^{-}(a)$ are finite set of ground $Class/2$ facts, and
$H = \{C\}$,
$AgreePos(H)$ = $\{Class(a',c): Class(a',c) \in (E^{+}(a) \cap MM(B \cup \{C\} \cup InvF)$
and
$AgreeNeg(H)$ = $\{Class(a',c): Class(a',c) \in (E^{-}(a) - (E^{-}(a) \cap MM(B \cup \{C\} \cup InvF)$.
From Remark \ref{rem:struc_unstruc}, $MM(B \cup \{C\} \cup InvF) = MM(B \cup \{C'\} \cup InvF)$. Since $H' = \{C\} \cup InvF$, it follows immediately that
$AgreePos(H) = AgreePos(H')$ and $AgreeNeg(H) = AgreeNeg(H')$ and therefore
$Fidelity(H|D,B) = Fidelity(H'|D,B)$.
\hfill $\Box$
\end{myremark}

\section{Selecting a Local Explanation}
\label{sec:sel}

Given a relational instance
$a \in {\cal X}$, let the prediction
of $a$ by a DRM $N$ be $c$. But, for
a definition of a neighbourhood, there
may be several explanations for $Class(a,c)$ with the same
maximal fidelity.
How then should we select a
single explanation?
Provided we have some reasonable way
of specifying prior-preferences, Bayes rule trades-off the fit to the data (likelihood)
against prior preference (LIME's minimisation of the sum of a loss and a regularisation
term can be seen as implementing a form
of Bayesian selection~\cite{lime}).

A general setting for selecting amongst logical formulae is provided
by labelled deductive systems (LDS: see \cite{Gabb:b:1996}), in which logical
formulae are extended with labels, with an associated algebra. 
For our purpose, it is sufficient simply to consider the comparison of labelled
explanations.

\begin{mydefinition}
{\bf Labelled Explanation.}
Given a relational example $e = Class(a,c)$, and background knowledge $B$,
$\alpha:H$ is a labelled explanation for $e$ given $B$ if: (a)
$H$ is an explanation for $e$ given $B$; and (b) $\alpha$ is a element
of a partially-ordered set of ground first-order terms $\Delta$. $\Delta$ consists of annotations of all explanations for $e$ given $B$.

\end{mydefinition}

A comparison of labelled explanations follows simply from the partial ordering
on the labels. That is, $\alpha:H_1 \preceq \beta:H_2$ iff $\alpha \preceq \beta$.
Here we will take the label of an explanation $H$ to be a pair $\langle L_H, P_H \rangle$,
which allows several different kinds of comparisons, given background knowledge $B$
and data $D$ (see Definition~\ref{def:fidelity}):

\begin{description}
    \item[Quantitative.] This is appropriate when both $L_H$ and $P_H$
    	are on an interval scale (that is, numeric values). Examples are: (a) the 
    	usual Bayesian comparison, using $L_H = P(D|H,B)$ and $P_H = P(H|B)$;
        and $\langle L_{H_1},P_{H_1} \rangle:H_1 ~\preceq~ \langle L_{H_2},P_{H_2} \rangle:H_2$
        iff $\mathrm{log}L_{H_1} + \mathrm{log}P_{H_1}$
        $\leq$ $\mathrm{log}L_{H_2} + \mathrm{log}P_{H_2}$; (b) Good's 
        explicativity (Chapter 23 of~\cite{good:expl}, which uses the same 			$L_H, P_H$, but uses the function
        $\mathrm{log}L_H + \gamma \mathrm{log}P_H$ with
        $0 < \gamma < 1$; and (c) Likelihood-based, using  $L_H$ = $P(D|H,B)$
        	and $P_H$ is the uniform distribution.
   \item[Qualitative.] Here both $L_H$ and $P_H$ are both
   		on an ordinal scale (that is, only comparisons of values are
        possible). Examples are: (a) The
   		qualitative Bayesian comparison in the manner
   		proposed by \cite{coletti:bayes}. With some abuse of notation,
        $\langle L_{H_1},P_{H_1} \rangle:H_1 ~\preceq~ \langle L_{H_1},P_{H_1} \rangle:H_2$
        iff $L_{H_1} \preceq L_{H_2}$ and $P_{H_1} \preceq P_{H_2}$.
        If it is not the case that
        $\langle L_{H_1},P_{H_1} \rangle:H_1 ~\preceq~ \langle L_{H_2},P_{H_2} \rangle:H_2$
        or
        $\langle L_{H_2},P_{H_2} \rangle:H_2 ~\preceq~ \langle L_{H_1},P_{H_1} \rangle:H_1$,
        then the labelled explanations are not comparable; and
        (b) A dictionary-ordering, in which
        	$\langle L_{H_1},P_{H_1} \rangle:H_1 ~\preceq~ \langle L_{H_1},P_{H_1} \rangle:H_2$
        	iff $L_{H_1} \prec L_{H_2}$, or $L_{H_1} = L_{H_2}$ and $P_{H_1} \preceq P_{H_2}$.\footnote{
            This may not yield the same results as the qualitative Bayesian comparison
            above. Differences arise when
            $L_{H_1} \prec L_{H_2}$ but $P_{H_2} \preceq P_{H_1}$. Under the 						
            dictionary-ordering $H_2$ would be preferred, but
            the qualitative Bayesian approach would find
            $H_{1}$ and $H_{2}$ incomparable.}
	\item[Semi-Quantitative.] Here, one of $L_H$ or $P_H$ is on an
    	interval scale, and the other is on an ordinal scale.
       	The qualitative comparisons above can be
        adapted to this, by replacing $\preceq$ and $\prec$ with
        $\leq$ and $<$ for the numeric quantity.
\end{description}

In this paper, we will use the semi-quantitative dictionary ordering with
$L_H = Fidelity(H|D,B)$ and $P_H$ is 
an ordinal-valued prior based on an assessment of relevance.
Using the semi-quantitative setting and the dictionary ordering has
some advantages:

\begin{itemize}
    \item[(a)] Quantitative selection based on a Bayesian score
    	requires a definition of both $P(D|H,B)$ and $P(H|B)$. While
    	the first can be obtained easily enough, it is not obvious          
    	how to specify a prior distribution over explanations.
        The usual approach of using a mathematically convenient
    	function like $2^{-|H|}$, where $|H|$ is some measure of the
    	size of $H$, may not be appropriate translation of
        prior assessment of relevance of explanations; and
    \item[(b)] A qualitative Bayesian approach 
    	as defined above usually ends up with many incomparable explanations.
        The dictionary ordering decomposes the task
        of identifying explanations into two parts:
        the first part that maximises fidelity, and the
        second part that maximises the prior amongst maximal fidelity
        explanations. Under some circumstances (see the Appendix),
        maximising fidelity is equivalent to maximising log
        likelihood. In those cases, the dictionary ordering
        uses the prior to select amongst maximum likelihood
        explanations. Usefully, there is also an implementation
        benefit that follows from the result in Remark \ref{rem:struct_fidelity}:
        since structuring does not alter the fidelity, the first part
        can simply examine unstructured explanations. 
\end{itemize}

We turn now to prior information $P_H$ that captures some aspects
of what constitutes a comprehensible explanation.

\subsection{A Relevance-Based Prior}
\label{sec:prior}

In \cite{ashking:relev} the authors investigate the utility of including an
expert assessment of
the relevance of relations included in the background knowledge. 

\begin{myexample}{
\bf Relevance information in the trains problem.}
Suppose the background knowledge $B$ for the trains problem
contains definitions of predicates like $Has\_Car/2$,
$Short/1$, $Closed/1$, $Wheels/2$ and $Load/2$.
For the problem of classifying trains as east-bound
or west-bound, let us assume we are also given
domain knowledge in the form of the relevance
level of (sets) of predicates as follows:
$r_1: \{Wheels/2,Load/2\}, r_2: \{Short/1,Closed/1\}$,
and $r_1 \prec r_2$. That is, $Wheels$ and $Load$
are less relevant to the problem than $Short$ and $Closed$.
\end{myexample}

We note that this is different to the notion of logical relevance of features (defined
in terms of the entailment relation, $\models$). Here,
we are concerned with domain-relevance of predicates used to define those features.
This latter form of domain-specific relevance information can also
form the basis of a preference ordering over explanations.
Here is the view of the domain expert involved
in \cite{ashking:relev}:\footnote{R.D. King: personal communication}


\begin{quote}
{\small{
\it{ 
I think it is reasonable to argue that [a hypothesis using] more relevant prior background knowledge
is more probable. I think that what makes hypotheses more probable is also a function of whether
the predicates used are related to each other.  Often you see hypotheses that seem to mix apples
and oranges together, which seems to make little sense.  Though of course this mixing of predicates may be
because the ML system is trying to express something not easily expressible with the given background
predicates.}
}}
\end{quote}

\noindent
This suggests that relevance information can constitute an important source of
prior knowledge. One route by which this is exploited by ILP systems is in the form
of search constraints (``hypotheses that do not contain oxygens connected to hetero-aromatic rings
are irrelevant''), or, as in the case of \cite{ashking:relev}, in the incremental construction of
hypotheses. Our interest here is to extend this use of relevance to \emph{selection amongst hypotheses},
by devising a relevance-based partial ordering over hypotheses.\footnote{We will often
reuse the generic symbols
$\preceq$ and $\prec$ to denote partial- and total-orderings. The context will make it
clear which sets these relations refer to.}

\begin{mydefinition}{\bf Relevance-assignments and orderings.}
We assume that for some set of predicates ${\cal P}$ in background knowledge $B$, we have
domain-specific relevance labels drawn from a set ${\cal R}$.
A relevance assignment is a function $R:{\cal P} \rightarrow {\cal R}$.
We assume that there is domain-knowledge in the form of a total ordering $\prec_r$ over the
elements of ${\cal R}$. Then, for $a,b \in {\cal R}$, $a \preceq_r b$ iff $a \prec_r b$ or $a = b$.
We will call $\prec_r$ a relevance ordering.
\end{mydefinition}

\noindent
A relevance ordering naturally
results in the concept of ordered intervals: $[a,b]$ is an ordered relevance-interval
(or simply, a relevance-interval) if $a,b \in {\cal R}$ and $a \preceq_r b$.
It is not hard to see that with a finite set of relevance labels ${\cal R}$,
the set of relevance intervals is partially ordered. That is,
$[a,b] \preceq [c,d]$ iff $a \preceq_r c$ and $b \preceq_r d$.
In fact, the following slightly more general ordering will be more
useful for us.
%
\begin{mydefinition}{\bf Ordering over sets of relevance-intervals}\label{def:relorder}
Let $S$ and $S^\prime$ be sets of relevance intervals. Then
$S \preceq_i  S^\prime$ iff for every interval $[a,b]$ in $S$ there exists
at least one interval $[c,d]$ in $S^\prime$ s.t. $[a,b] \preceq [c,d]$.
That is, $a \preceq_r c$ and $b \preceq_r d$.
\end{mydefinition}

\noindent
We now construct, in stages, the relevance of an explanation. 

\begin{mydefinition}{\bf Relevance of features.}
Given background knowledge $B$, let $\prec_r$ be a relevance ordering
over a set of relevance labels ${\cal R}$, and let $R:{\cal P} \rightarrow {\cal R}$
be a relevance assignment for some subset ${\cal P}$ of $B$.
Let $\forall x(F(x) \leftarrow {Cp}(x))$ be the feature-definition for $F/1$ 
in which ${Cp}(x)$ is a conjunction containing
predicates from ${\cal P}$ only. Then
the relevance of the feature $F$ is $Relev(F) = [l,h]$, where $l$
is the minimum relevance of predicates in ${Cp}(x)$ according to $\prec_r$ and $R$, and
$h$ is maximum relevance of predicates in ${Cp}(x)$ according to $\prec_r$ and $R$.
\end{mydefinition}

\begin{mydefinition}{\bf Relevance of feature-clauses.}
Let ${\cal F}$ be a set of features.
Let $C$ be a feature-clause. W.l.o.g. let the features in $C$ be $\{F_1,F_2,\ldots,F_k\}$
where the $F_i \in {\cal F}$. Let $Relev(F_i) = [l_i,h_i]$.
Then $Relev(C) =\{[l*,h*]\}$ where $l* = \mathrm{min}(l_1,l_2,\ldots,l_k)$ and
$h* = \mathrm{max}(h_1,h_2,\ldots,h_k)$.
\end{mydefinition}

The need to have $Relev(C)$ as a set will become apparent shortly. The relevance of explanations is constructed from the relevance of the
feature-clauses in the explanation.

\begin{mydefinition}{\bf Relevance of an explanation} \label{def:relexp2}
Given background knowledge $B$, let $H$ be an explanation for $Class(a,c)$
containing the clause $C: \forall x(Class(x,c) \leftarrow Body)$.
If $H$ is an unstructured explanation then $Relev(H) = Relev(C)$. 
Otherwise, if $H$ is a structured explanation, then 
$Relev(H) = \bigcup_{F_i \in Body} Relev(UFC(F_i))$.
\end{mydefinition}

\noindent
It is interesting that although a pair of structured explanations
may unfold to the same unstructured explanation (and therefore
have the same fidelity), their relevance may not be the same.
Intuitively, structuring any unstructured explanation will split the
relevance-interval of the corresponding feature-clause into a set of intervals (see Definition~\ref{def:relexp2}).
In a ``good structuring'' each interval in this set will be 
``narrower'' than the unstructured relevance-interval and will therefore
be preferred under the relevance ordering (Definition~\ref{def:relorder}).

\begin{myexample}
{\bf Comparing relevance of explanations in the trains problem.}\label{ex:relexp}
Suppose we are given a set of features ${\cal F} = \{F_2,F_3,F_4,F_9\}$ and
following feature-definitions (omitting quantifiers for simplicity):\\

\noindent
${F}_2(x) \leftarrow Has\_Car(x,y), Short(y)$\\
${F}_3(x) \leftarrow Has\_Car(x,y), Closed(y)$ \\
${F}_4(x) \leftarrow Has\_Car(x,y), Wheels(y,3)$ \\
${F}_9(x) \leftarrow Has\_Car(x,y), Load(y,triangle)$ \\

\noindent
Let as assume we are given a set of relevance labels ${\cal R}$ = $\{r1,r2\}$,
with $r1 \prec r2$. Let us further assume the following
relevance-assignment: $\{(Wheels/1,r1),(Load/2,r1),$ $(Short/1,r2),(Closed/1,r2)\}$.
That is $Relev(F_2) = Relev(F_3) = [r2,r2]$, $Relev(F_4) = Relev(F_9) = [r1,r1]$.
Suppose we have  the structured explanation $H_1$: \\

{\small{
\begin{tabular}{l} \\
$Class(x,East) \leftarrow {F}_{1,1}(x), {F}_{1,2}(x)$\\
${F}_{1,1}(x) \leftarrow {F}_{2}(x), {F}_{3}(x)$  \\
${F}_{1,2}(x) \leftarrow {F}_4(x), {F}_9(x)$\\
\end{tabular}
}}

\vspace*{0.2cm}
\noindent
Clearly, $F_{1,1}, F_{1,2} \not \in {\cal F}$, and are therefore ``invented'' features.
Let ${\cal F}_2$ = $\{F_{1,1}, F_{1,2}\}$.
Then, $Relev(H_1) = R_{H_1} = \bigcup_{F \in {\cal F}_2} Relev(FC(F)) \cup \emptyset$.
Now $Relev(FC(F_{1,1}))$ = $\{[r2,r2]\}$,
$Relev(FC(F_{1,2}))$ = $\{[r1,r1]\}$ and $R_{H_1} = \{[r1,r1],[r2,r2]\}$.

%

On the other hand, for the following explanation $H_2$:\\

{\small{
\begin{tabular}{l} \\
$Class(x,East) \leftarrow {F}_{1,3}(x), {F}_{1,4}$\\
${F}_{1,2}(x) \leftarrow {F}_{3}(x), {F}_{4}(x)$  \\
${F}_{1,4}(x) \leftarrow {F}_{2}(x), {F}_{9}(x)$\\
\end{tabular}
}}

\vspace*{0.2cm}
\noindent
$R_{H_2} = \{[r1,r2]\}$. From Defn. \ref{def:relorder}
$R_{H_2} \preceq_i R_{H_1}$. Note: $H_1$ and $H_2$ both unfold to
the unstructured explanation $H$: $Class(x,East) \leftarrow {F}_{2}(x),$ ${F}_{3}(x), {F}_{4}(x), {F}_{9}(x)$
for which $R_{H} = \{[r1,r2]\}$.

Thus, although $H_{1,2}$ both unfold to $H$, it is possible that a selection criterion that
takes relevance into account may prefer $H_1$ over $H_2$ and $H$.
\end{myexample}


\begin{myremark}
{\bf Structuring can increase relevance.} \label{rem:struc_relev}
Let $H = \{C\}$ be an unstructured explanation for
$Class(a,c)$, and let $H'$ be a structured
explanation containing a clause
$C': Class(x,c) \leftarrow Body$
that unfolds to $C$. Let
$P_H = Relev(H)$ and $P_{H'} = Relev(H')$. Then $P_H \preceq P_{H'}$. \\
Let $C$ contain the features $\{F_1,\ldots,F_l\}$, where $Relev(F_i) = [l_i,h_i]$
Since $H$ is an unstructured explanation, $Relev(H)$ $=$ $\{[l*,h*]\}$, where
$l* = \mathrm{min}(l_1,\ldots,l_k)$ and $h* = \mathrm{max}(h_1,\ldots,h_k)$.
Let $C'$ contain the invented features $\{{F'}_1,\ldots,{F'}_j\}$. 
Since  $C'$ unfolds to $C$, each ${F'}_i$ unfolds to a clause containing
some subset ${S'}_i$ of $\{F_1,\ldots,F_k\}$,
and $\bigcup_{i=1}^j {S'}_i ~=~ \{F_1,\ldots,F_k\}$.
W.l.o.g. let $h* = h_k$. By the constraint imposed on structured explanations,
there must be at least one invented feature ${F'}_m$ that
unfolds to a clause containing $F_k$. Let $Relev(UFC({F'}_m)) = [l'_m,h'_m]$.
Clearly, $l* \preceq l'_m$ and $h* = h'_m$, and therefore $[l*,h*] \preceq [l'_m,h'_m]$.
Since $[l'_m,h'_m] \in Relev(H')$, it follows from Defn. \ref{def:relorder}
that $Relev(H) \preceq Relev(H')$.
\end{myremark}

\subsection{Implementation}
\label{sec:implement}

We finally have the pieces to
define a label for an explanation: each explanation
$H$ will now have the label $\langle L_H, P_H \rangle$,
where $L_H = Fidelity(H|D,B)$ and $P_H = Relev(H)$.
Using a dictionary-ordering to compare labelled explanations
allows us to decompose the task of identifying explanations
into two parts: the first that maximises fidelity and the second that maximises the
relevance. Further, as we have already seen (Remark \ref{rem:struct_fidelity}),
structuring cannot increase fidelity, but can increase relevance (Remark \ref{rem:struc_relev}).     
  Therefore, with a dictionary ordering on labels,
  it suffices to search first over the space of unstructured explanations,
  and then over the space of structured explanations that unfold to the unstructured explanations
  with maximal fidelity. Figure ~\ref{fig:unstruct_optim} extends the
  previous procedure of finding an unstructured explanation (Fig.~\ref{fig:unstruct})
  to obtain the highest-fidelity unstructured explanation.

\begin{figure*}[htb]
{\small{
$ConstructUnstruct(e,B,{\cal F},E^{+},E^{-}):$
    \begin{description}
    \item[Given:] A relational example $e$; 
        background knowledge $B$;
        a set of features ${\cal F}$ with definitions in $B$; and $E^{+}, E^{-}$ as defined
        in Defn. \ref{def:local}.
    \item[Find:] A maximal fidelity unstructured explanation
    		$H$ s.t. $B \cup H \models e$,
    \end{description}
\begin{enumerate}
	\item Let $e = Class(a,c)$
    \item Let $a' = FV(a)$
     \item Let ${\cal F}'$ be the set of features
        that map to $TRUE$ in $a'$
    \item Let $D = (E^{+},E^{-})$
    \item Let ${\cal H}$ be the subset-lattice of ${\cal F}'$
    \item Let $F'$ be any element in ${\cal H}$ s.t.
            \begin{itemize}
                \item $Body$ is the conjunction of features in $F'$;
                \item $C = \forall x(Class(x,c) \leftarrow Body)$;
                \item $L = Fidelity(\{C\}|D,B)$; and
               \item There is no other element $F''$ in ${\cal H}$ s.t. $Fidelity(\{Clause(a,F'')\}|D,B) > L$.
            \end{itemize}
    \item return $\{C\}$
\end{enumerate}
}}
\caption{A procedure for identifying an unstructured explanation with maximal fidelity. In practice, we will need to extend this procedure to
    return all unstructured
    explanations with maximal fidelity.}
\label{fig:unstruct_optim}
\end{figure*}

Figure \ref{fig:struct_optim} extends $ConstructStruct$ in Fig.~\ref{fig:struct} to
return an explanation with higher-relevance than an unstructured $H$, if one exists.
It is not hard to see that if $P_k = \emptyset$ then $P_{k+1} = \emptyset$.
Therefore, it is only needed to call $ConstructExpl$
with $k=2,3,\ldots$ until $P_k = \emptyset$. In experiments in this
paper, we will adopt the even simpler strategy of only considering $k=2$. That
is, we will only consider 2-partitions of the set of features constituting
the unstructured explanation $H$ (in effect, seeking structured explanations
with higher relevance than $H$, but using the minimum number of invented features). The structured explanations in Example
\ref{ex:strucexptrains} are examples of structures that
can be obtained with $k=2$.

\begin{figure*}[htb]
{\small{
$ConstructExpl(H,B,k):$
    \begin{description}
        \item[Given:] An unstructured explanation $H$; background knowledge $B$; and a number $k$ ($\geq 2$)
        \item[Find:] An explanation $H'$ s.t. $Relev(H|B) \preceq Relev(H'|B)$.
    \end{description}
\begin{enumerate}
\item Let $H$ = $\forall x(Class(x,c) \leftarrow Body)$ 
\item Let $Relev(H) = R_H = \{[\alpha,\gamma]\}$
\item Let $F$ be the set of features in $Body$
\item Let $P_k$ be a set s.t:
	\begin{enumerate}
		\item[(a)] Either $P_k$ is a $k$-partition of $F$ that satisfies:
    	\begin{itemize}
    		\item At least one block in $P_k$
        			contains $2$ or more elements; and
    		\item There is at least one block in $P_k$ whose elements have
        		a minimum relevance $\beta$ and maximum relevance $\gamma$
        		such that $\alpha \prec_r \beta \preceq_r \gamma$
  		\end{itemize}
 		\item[(b)] Or, $P_k = \emptyset$ (if no such $k$-partition exists)
 	\end{enumerate}
\item If $P_k = \emptyset$ return $H$. \label{step:part}
\item Otherwise:
\begin{enumerate}
    \item Let $P_k$ consist of blocks $b_1,b_2,\ldots,b_k$
    \item For each block $b_i \in P_k$:
    \begin{enumerate}
        \item Construct a new feature ${F'}_i$ with
            definition $C_i$ = $\forall x({F'}_i(x) \leftarrow {Body}_i)$,
            where ${Body}_i$ is the conjunction of the features in $b_i$
    \end{enumerate}
    \item Let $C_0$ = $\forall x(Class(x,c) \leftarrow {Body}_0)$ where
            ${Body}_0$ is the conjunction of the ${F'}_1,{F'}_2,\ldots,{F'}_k$
    \item Let $H'$ = $\bigcup_{i=0}^k C_k$
\end{enumerate}
\item Return $H'$.
\end{enumerate}
}}
\caption{A procedure for obtaining a structured explanation that is at
	least as relevant as an unstructured explanation $H$. 
	The structured explanation is obtained by inventing
    $k$ features, the definition of at least one of which has a higher relevance than
    the unstructured explanation. }
\label{fig:struct_optim}
\end{figure*}

Together, $ConstructUnstruct$ and $ConstructExpl$ are used
to identify a local explanation for a relational instance $e$ (Fig.\ref{fig:smallpic2}).

\begin{figure}

\centerline{\includegraphics[height=0.36\textheight,width=0.8\textwidth]{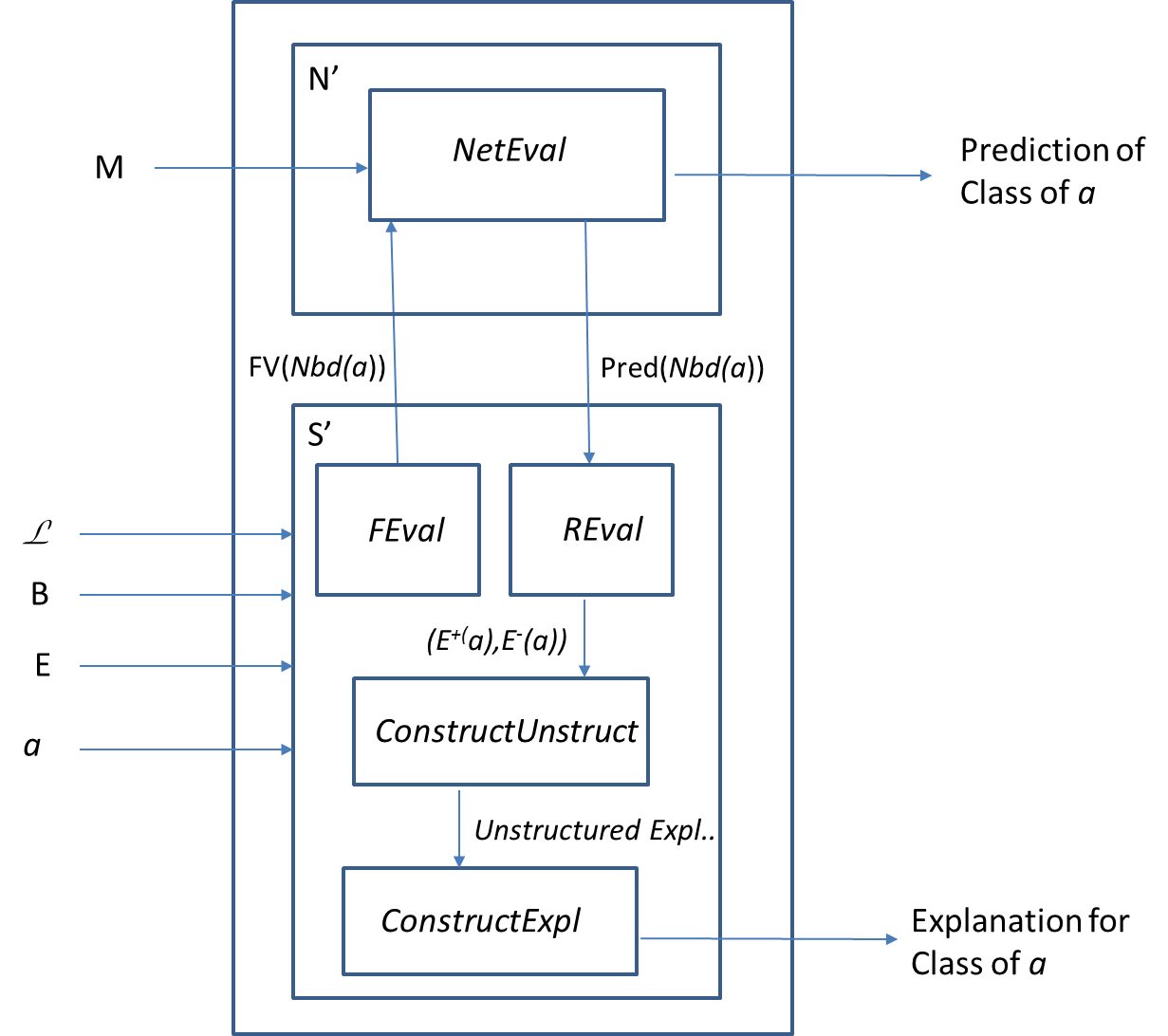}}
\caption{More details on the testing component in Fig.~\ref{fig:bigpic}. 
$E$ is the set of relational training instances, $a$ is a test instance,
and $M$ is a DRM.
FV($Nbd(a)$) denotes the feature-vectors for relational in $E$ that are within
the neighbourhood of $a$. Pred($Nbd(a)$) are the class predictions of instances in $Nbd(a)$ by
the DRM. $NetEval$ evaluates the model $M$ on input data; $FEval$ evaluates relational instances and returns their feature-vectors; and $REval$ constructs relational examples. We assume that $FEval$ has access to the feature-definitions found during the training stage.}
\label{fig:smallpic2}
\end{figure}
\section{Empirical Evaluation}
\label{sec:expt}

In this section, we evaluate empirically the predictive performance of
DRMs and the explanatory models derived from them.
%
%
Our aim is investigate the following:
\begin{description}
    \item[Prediction.] We conduct the following experiment:
        \begin{description}
            \item[Expt. 1: Accuracy.] Will a DRM constructed using randomly
                drawn features from a depth-limited mode language have good predictive
                performance?
           \end{description}
    \item[Explanation.] We conduct the following experiments:
            \begin{description}
                \item[Expt. 2: Fidelity.] Can we construct a local symbolic
                    explanations for an instance with high fidelity to local
                    predictions made by the DRM?
                \item[Expt. 3: Relevance.] Does incorporating s prior preference
                    based on relevance have any effect?
            \end{description}
\end{description}

\noindent
Some clarifications are necessary here: (a) By randomly drawn features in Expt. 1, we
mean the rejection-sampling method described in Section \ref{sec:sample}; (2) By a local symbolic
explanation in Expt. 2 we mean the use of a graph-search that returns the unstructured
explanation with the highest fidelity, described in Section~\ref{sec:localexp}; and (3)
By prior-preference in Expt. 3, we mean the relevance-based ordering over structured
or unstructured explanations as defined in Defn. \ref{def:relexp2}. In Expt. 3, we
confine ourselves to whether the use of the preference can change the explanation
returned (either from a unstructured to a structured one, or from one unstructured explanation
to another).  We note that incorporation of prior preference obtained from a
human expert is still not sufficient to ensure
comprehensibility of explanations by the expert.
Evidence for this requires results in the form of cross-comparisons on
the use of prior expert preference on explanations
against expert comprehensibility of explanations. However, this is outside the
scope of this paper.

\subsection{Materials}

\subsubsection{Data}
We report results from experiments conducted using $7$
 well-studied real world problems from the ILP literature. These are:
Mutagenesis \cite{King_96}; Carcinogenesis \cite{King_96a}; DssTox \cite{Muggleton_08a};
and 4 datasets arising from the comparison of Alzheimer's drugs
denoted here as $Amine$, $Choline$, $Scop$ and $Toxic$ \cite{Srinivasan_1996y}.
Each of these have shown to benefit from the use of a first-order
representation, and domain-knowledge but there is still room for
improvement in predictive accuracies. Importantly, for each dataset, 
we also have access to domain-information about the relevance of
predicates for the classification task considered.

%
%
Of these datasets, the first three (Mut188--DssTox) are predominantly relational in nature,
with data in the form of the 2-d structure of the molecules (the
atom and bond structure), which can be of varying sizes, and diverse.
Some additional bulk properties of entire molecules
obtained or estimated from this structure are also available.
The Alzheimer datasets (Amine--Toxic) are best thought of
as being quasi-relational. The molecules have a fixed template, but vary in
number and kinds of substitutions made for positions on the template. A first-order
representation has still been found to be useful, since it allows expressing
concepts about the existence of one or more substitutions and their properties.
The datasets range in size from a few hundred (relational) instances to a few thousands. This
is extremely modest by the usual data requirements for deep learning. We refer the reader to the references cited for details
of the domain-knowledge used for each problem.

\subsubsection{Background Knowledge}

For the relational datasets (Mut188--DssTox), background knowledge is in the
form of general chemical knowledge of ring-structures and some functional groups.
Background-knowledge contains definitions used for concepts like: alcohols, aldehydes, halides, amides, amines,
acids. esters, ethers, imines, ketones, nitro groups, hydrogen donors and acceptors,
hydrophobic groups, positive- and negatively-charged groups,
aromatic rings and non-aromatic rings, hetero-rings, 5- and 6-carbon rings and so on.
These have been used in structure-activity applications
of ILP before \cite{king:mut,ashking:ehp}. 
However, we note that none of these definitions are specifically designed for the
tasks here. In addition, for Mut188 and Canc330,
there are some bulk properties of the molecules that are available. For the Alzheimer problems
(Amine--Toxic) domain knowledge consists of properties of the substituents in terms of
some standard chemical measures like size, polarity, number of hydrogen donors and acceptors and so on.
Predicates are also available to compare these values across substitutions. Again we refer
the reader to the relevant ILP literature for more details.

In addition to the domain-predicates just described, we will also
have information in the form of a relevance
ordering as described in \cite{ashking:relev}. That paper only refers to the Mut188 and Canc330
datasets. The same information is obtained from the domain-expert involved in that paper
for the other problems in this paper. A complete description of the relevance
assignment of predicates for each problem is in Appendix \ref{appa}

\subsubsection{Algorithms and Machines}

Random features were constructed on an Intel Core i7 laptop computer, using
VMware virtual machine running Fedora 13, with an allocation of 2GB for
the virtual machine. The Prolog compiler used was Yap. Feature-construction
uses the utilities provided by the Aleph ILP system \cite{aleph} for
constructing most-specific clauses in a depth-bounded mode language, and
for drawing clauses subsuming such most-specific clauses. No use is made
of any of the search procedures within Aleph.
The deep networks were constructed using the Keras library with 
Theano as the backend, and were trained using an NVIDIA K-40 GPU card. 

\subsection{Methods}

The methods used for each of the experiments are straightforward

\begin{description}

\item[Prediction.] Experiment 1 is concerned solely with the predictive
    performance of the DRM.
\begin{itemize}
\item[] For each dataset:
    \begin{enumerate}
        \item Obtain a set of random features ${\cal F}$;
        \item Compute the Boolean-value for each $F \in {\cal F}$ for the data;
        \item Construct a DRM $N$ using training data and obtain its predictions on
                test instances;
        \item Estimate the overall predictive performance of $M$ (Expt. 1)
    \end{enumerate}
\end{itemize}
\item[Explanation.] Experiments 2 and 3 are concerned solely with the explanatory
    performance of the local symbolic models.
\begin{itemize}
\item[] For each dataset:
    \begin{enumerate}
        \item Construct a DRM using training data
        \item For each test instance, obtain the local symbolic unstructured explanation(s) with
            the highest fidelity;
        \item Estimate the overall fidelity of the symbolic explanations (Expt. 2)
        \item Estimate the effect of using the relevance-based prior in hypothesis selection (Expt. 3)
    \end{enumerate}
\end{itemize}
\end{description}

\noindent
Some clarifications are necessary at this point:
\begin{itemize}
\item We use a straightforward  Deep Neural Network (DNN) architecture.
    There are multiple, fully connected feedforward layers of rectified linear (ReLU) units 
    followed by Dropout for regularization (see \cite{bengio:dl} for a description of these ideas). 
    The model weights were initialized with a Gaussian distribution.
    The number of layers, number of units for each layer, the optimizers, and other training hyperparameters
    such as learning rate, were determined via a validation set, which is part of the training data.
    Since the data is limited for the datasets under consideration, after obtaining the model
    which yields the best validation score, the chosen model is then
    retrained on the complete training set (this includes the validation set) until the training
    loss exceeds the training loss obtained for the chosen model during validation.
   \item We use the Subtle algorithm \cite{subtle} to perform
    the subsumption-equivalence test used to determine redundant features.
    \item For all the datasets, 10-fold cross-validated estimates of the 
    	predictive performance using ILP methods are available
        in the ILP literature for comparison. We use the same approach. 
        This requires constructing DRMs separately for each of the
        cross-validation training sets, and testing them on the
        corresponding test sets to obtain estimates of the predictive accuracy;
    \item We use the mode-language and depth constraints for the datasets
        that have been used previously in the ILP literature. For the
        construction of features, the rejection-sampler performs at most $10,000$ draws;
   \item We take \emph{explanatory fidelity} to mean the probability that
        the prediction made by $H$ on a randomly
        drawn instance agrees with the prediction made by the corresponding DRM.
        We use the same 10-fold cross-validation strategy for estimating this probability (for efficiency,
        we use the same splits as those used to estimate predictive accuracy).
        For a given train-test split ${Tr}_i$ and ${Te}_i$, we proceed as follows. We obtain a DRM $N_i$ using ${Tr}_i$.
        We start with a fidelity count of 0. For each instance ${x'}_j$ in ${Te}_i$ we obtain
        the class predicted by $N_i$ for ${x'}_j$, and
        corresponding neighbourhood of ${x'}_j$ in the training set ${Tr}_i$.
        The neighbourhood is partitioned into $\delta^{+}({x'}_j)$ and $\delta^{-}({x'}_j)$
        using the predictions by $N_i$ and high-fidelity
        unstructured explanation(s) $H_{ij}$ are obtained.
        This is done using a beam-search over the lattice
        described in Figs. \ref{fig:unstruct}, \ref{fig:unstruct_optim}.
        The size of the beam is 5 (that is, the top 5 unstructured
        explanations are returned).
    \item Assessments of fidelity require the definition of
        a neigbourhood. For each instance ${x'}_j$ in ${Te}_i$ we say any instance $x \in {Tr}_i$
        is in the neighbourhood of $x'$ iff $FV(x')$ and $FV(x)$ differ in no more
        than $k$ features. This is just the Hamming distance between
        the pair of Boolean vectors.
        That is, the neighbourhood of
        a test instance $x'$ consists of training instances $x$'s whose feature-vector
        representation are within a $k$-bit Hamming distance of $x'$.
        We will consider $k=5$ and $k=10$ in the experiments.
    \item In all cases, estimates are obtained using the same 10-fold cross-validation
        splits reported in the ILP literature
        for the datasets used. This will
        allow a cross-comparison of the results from Expt.1 to those reports.
        
\end{itemize}

\subsection{Results}
\label{sec:results}

Results of the empirical evaluation are tabulated in Fig.~\ref{fig:resultspred},
and Fig.~\ref{fig:resultsexpl}.
Some supplementary results are in Fig.~\ref{fig:results2}.
The principal observations that can be
made from the main tabulations in Figs.~\ref{fig:resultspred},\ref{fig:resultsexpl} are these:
(1) The predictive accuracy of the DRMs clearly compare favourably to the best
    reports in the literature;
(2) High-fidelity symbolic explanations can be obtained for local predictions
    made by the DRM; and
(3) In $6$ of the $7$ problems, introducing a prior-preference based on relevance does affect
    the selection of explanations.

\noindent
Together, the results provide evidence for the following:
 \begin{enumerate}
\item[(a)] A deep relational machine (DRM) equipped with domain knowledge and
    randomly drawn first-order features can construct good predictive
    models using (by deep-learning standards) very few data instances;
 \item[(b)] It is possible to extract symbolic explanations for the prediction
    made by the DRM for a randomly drawn (new) instance. The explanations
    are largely consistent with the predictions of the
    network for that instance and its near-neighbours in the training data
    examined before by the network; and
\item[(c)] It is possible to
incorporate domain-knowledge in the form
    of expert assessment of relevance of background predicates into
    a preference ordering for selecting amongst explanations.
\end{enumerate}

Quantitative assessments of the results are also possible, with the usual cautions associated
with small numbers and multiple comparisons. The appropriate test for comparing the predictive
accuracy of the DRM is the Wilcoxon signed-rank test, with the null hypothesis that the DRM's
accuracy is the same as the method being compared. This yields $P$-values of $< 0.05$ for
the comparisons against $OptILP$ and $Stat$ (we omit a comparison against the DRM in \cite{lodhi:drm},
due to lack of data). If the inclusion of relevance does not make a difference to selecting an explanation, we would expect
values of $0.0$ in Fig. \ref{fig:resultsexpl}(b). It is evident that the observed values are clearly not $0$, for
all cases except $DssTox$. The exception is unsurprising, since all predicates used for this problem
have the same relevance (see Appendix \ref{appa}). 

The results obtained are presented in some more detail in Fig.~\ref{fig:results2}. From this we observe:

\begin{itemize}
\item[(a)] The neighbourhood size affects the local explanations constructed. In general,
    the fewer the instances in the local neighbourhood, the lesser the constraints
    imposed on the explanations. This leads to smaller explanations (fewer literals), and higher fidelity;
\item[(b)] Although the networks can often contain 1000's of input features (the exact numbers for
    each problem are not shown here, but range from about 2000 (Amine) to 7000 (Mut188)),
    a high-fidelity locally-consistent explanation may only contain a few (``active'') features.
    This is what makes it possible to extract relatively compact explanations even for large networks
    (recall that we are not attempting to extract a complete symbolic model for the entire network);
\item[(c)] There are clear differences between the role of the relevance information amongst the
    datasets (significant effect in Mut188 and Canc330; minor effect in the Alzheimer datasets; and
    no effect in DssTox). We conjecture the following condition for relevance to
    have an effect on selection:
        \begin{itemize}
            \item[] {\it The larger the range of the relevance
                assignment, the more likely it is that relevance will play a role in selection of explanations\/}.
        \end{itemize}
    For the datasets here, the range of the relevance assignment has 1 value for DssTox; 2 values for the Alzheimer's
    datasets; and 4 values each for Mut188 and Canc330. The corresponding proportions of explanations where
    relevance plays a role in selection are: 0.0 (DssTox); 0.29 (Alzheimer datasets); and 0.80 (Mut188 and Canc330).
\item[(d)] Structured explanations only appear to play role for larger neighbourhoods. We suggest this is not so
    much to do with the size of the neighbourhood, as to the corresponding increasing in the size of
    the explanations. In general, larger explanations (those with more features) are likely to benefit from structuring.
\end{itemize}

\begin{figure}
\begin{center}
{\small{
\begin{tabular}{|l|c|c|c|c|}\hline
         & \multicolumn{4}{|c|}{Accuracy}\\ \cline{2-5}
Problem & $OptILP$     & $Stat$   & $DRM$      & $DRM$ \\
         & \cite{ash:optilp} & \cite{Amrita12}& \cite{lodhi:drm} & (here) \\ \hline
         
$Mut188$  & 0.88(0.02) & 0.85(0.05)  & 0.90(0.06) & 0.91(0.06)\\
$Canc330$ & 0.58(0.03) & 0.60(0.02)  &  --        & 0.68(0.03)\\
$DssTox$  & 0.73(0.02) & 0.72(0.01)  & 0.66(0.02) & 0.70(06)\\
$Amine$   & 0.80(0.02) & 0.81(0.00)  & --         & 0.89(0.04)\\
$Choline$ & 0.77(0.01) & 0.74(0.00)  & --         & 0.81(0.03)\\
$Scop$    & 0.67(0.02) & 0.72(0.02)  & --         & 0.82(0.06)\\
$Toxic$   & 0.87(0.01) & 0.84(0.01)  & --         & 0.93(0.03) \\ \hline
\end{tabular}
}}
\end{center}
\caption{Experiment 1. Estimated predictive accuracies of DRMs against some
    of the best reported performances in the ILP literature. All
    estimates are from the same 10-fold cross-validation splits
    in the reports cited.}
\label{fig:resultspred}
\end{figure}

\begin{figure}
\begin{minipage}[h]{0.5\textwidth}
\begin{center}
{\small{
\begin{tabular}{|l|c|}\hline
Problem & {Fidelity} \\ \hline
$Mut18$   & 0.99(0.01) \\
$Canc330$ & 0.99(0.01) \\
$DssTox$  & 0.87(0.03)\\
$Amine$   & 0.98(0.01)\\
$Choline$ & 0.89(0.01)\\
$Scop$    & 0.89(0.02)\\
$Toxic$   & 0.94(0.02)\\ \hline
\end{tabular}
}}
\end{center}
\begin{center}
(a)
\end{center}
\end{minipage}
\begin{minipage}[h]{0.5\textwidth}
\begin{center}
{\small{
\begin{tabular}{|l|c|}\hline
Problem & $\langle L_H,R_H\rangle > \langle L_H,\emptyset \rangle$\\ \hline
$Mut188$  & 0.82(0.08)\\
$Canc330$ & 0.77(0.07)\\
$DssTox$  & 0.00(0.00) \\
$Amine$   & 0.38(0.15) \\
$Choline$ & 0.27(0.03) \\
$Scop$    & 0.27(0.06) \\
$Toxic$   & 0.24(0.14) \\ \hline
\end{tabular}
}}
\end{center}
\begin{center}
(b)
\end{center}
\end{minipage}
\caption{Experiments 2 and 3. (a) Mean fidelity of the explanatory (symbolic) model to the predictive (neural) model.
    The number tabulated is
    a 10-fold cross-validation estimate of the faithfullness of the symbolic model
    to a DRM's prediction assessed over the neighbourhood of a test instance.
    The entries are for the smallest neighbourhood ($H_5$:
    see the ``Methods'' section for how this is computed).
    The number in parentheses are estimates of standard deviations.
    (b) Relative frequency estimates of how often we can expect incorporation
        of a relevance-based prior to affect the selection of explanations. The tabulation
        is the proportion of explanations for which a
        Bayes label using both fidelity and relevance ($\langle L_H,R_H \rangle$) is better
        than one that uses fidelity only ($\langle L_H,\emptyset \rangle$). The explanations
        are for the $H_5$ neighbourhood. Again,
        the estimates are from the same 10-fold cross-validation splits used
        elsewhere.}
\label{fig:resultsexpl}
\end{figure}

\begin{figure}
\begin{minipage}[h]{0.5\textwidth}
\begin{center}
{\small{
\begin{tabular}{|l|cc|}\hline
Problem & \multicolumn{2}{|c|}{Nbd. Size} \\ \cline{2-3}
          & $H_{5}$  & $H_{10}$ \\ \hline  
$Mut188$  & 5(1) & 20(3) \\
$Canc330$ & 3(1) & 27(4) \\
$DssTox$  & 19(3) & 83(10)  \\
$Amine$   & 9(1) & 23(20 \\
$Choline$ & 14(1) & 53(3) \\
$Scop$    & 8(1)  & 26(2) \\
$Toxic$   & 12(2)& 42(2) \\ \hline
\end{tabular}
}}
\end{center}
\begin{center}
(a)
\end{center}
\end{minipage}
\begin{minipage}[h]{0.5\textwidth}
\begin{center}
{\small{
\begin{tabular}{|l|cc|}\hline
Problem & \multicolumn{2}{|c|}{Fidelity} \\ \cline{2-3}
        & $H_5$ & $H_{10}$  \\ \hline
$Mut188$  & 0.99(0.01)  & 0.96(0.01) \\
$Canc330$ & 0.99(0.01)  & 0.92(0.02) \\
$DssTox$  & 0.87(0.03)  & 0.85(0.02) \\
$Amine$   & 0.98(0.01)  & 0.97(0.01) \\
$Choline$ & 0.89(0.01)  & 0.83(0.01) \\
$Scop$    & 0.89(0.02)  & 0.86(0.02) \\
$Toxic$   & 0.94(0.02)  & 0.90(0.03)\\ \hline
\end{tabular}
}}
\end{center}
\begin{center}
(b)
\end{center}
\vspace*{0.2cm}
\end{minipage}
\begin{minipage}[h]{0.5\textwidth}
\begin{center}
{\small{
\begin{tabular}{|l|cc|}\hline
Problem & \multicolumn{2}{|c|}{Expl. Size} \\ \cline{2-3}
        & $H_5$ & $H_{10}$ \\ \hline
$Mut188$  & 1(1) & 2(1) \\
$Canc330$ & 1(1) & 3(1) \\
$DssTox$  & 3(1) & 4(1)  \\
$Amine$   & 2(1) & 2(1) \\
$Choline$ & 2(1) & 3(1) \\
$Scop$    & 2(1) & 2(1)   \\
$Toxic$   & 2(1) & 3(1) \\ \hline
\end{tabular}
}}
\end{center}
\begin{center}
(c)
\end{center}
\end{minipage}
\begin{minipage}[h]{0.5\textwidth}
\begin{center}
{\small{
\begin{tabular}{|l|cc|}\hline
Problem & \multicolumn{2}{|c|}{Struc. Expl.} \\ \cline{2-3}
        & $H_5$ & $H_{10}$  \\ \hline
$Mut188$  & 0.00 & 0.04 \\
$Canc330$ & 0.02 & 0.36 \\
$DssTox$  & 0.00 & 0.00 \\
$Amine$   & 0.00 & 0.00 \\
$Choline$ & 0.00 & 0 00 \\
$Scop$    & 0.00 & 0.00 \\
$Toxic$   & 0.00 & 0.00 \\ \hline
\end{tabular}
}}
\end{center}
\begin{center}
(d)
\end{center}
\end{minipage}
\caption{Effect of the neighbourhood.
        (a) Average numbers of neighbouring instances for local explanations (rounded up);
        (b) Average fidelity of local explanations;
        (c) Average number of literals in the maximal-fidelity explanations (rounded up); and
        (d) Average proportion of structured explanations.
        All numbers are
        estimates obtained from 10-fold cross-validation, and the number in parentheses is
        an estimate of the standard deviation.}
\label{fig:results2}
\end{figure}

Finally, since explanations are generated ``on-demand'', it is impractical to show the explanations for all test instances. A snapshot is nevertheless useful, and is
included in Appendix \ref{app:expl}. The example is from the $Canc330$ problem, and shows 4 possible explanations. The unstructured explanation shown has perfect fidelity. The 3 alternate structured explanations derived from this unstructured explanation
have the same fidelity (as expected), but higher relevance. In effect, what the
structuring achieves here is to group predicates with the $Has\_property$
relation, which has high-relevance; and separate out predicates that mix
predicates with low- and high-relevance. Some measure of the explanatory
convenience provided by the symbolic model is apparent if the reader keeps in mind
that the corresponding prediction by the DRM is based on around 2000
input features, about 400 of which are equal to $1$ for the test
instance. It is interesting that the domain-expert could correctly identify
the class of the example when shown the symbolic explanation.

\section{Other Related Work}
\label{sec:relwork}

We have already noted the key reference to LIME and to reports in the ILP literature of immediate relevance to
the work in this paper. Here we comment on other related work.
The landmark work on structured induction of symbolic models is that
of Shapiro \cite{shap:thesis}. There, structuring was top-down
with machine learning being used to learn sub-concepts
identified by a domain-expert. The structuring is therefore
hand-crafted, and with a sufficiently well-developed tool, a
domain-expert can, in principle, invoke a machine learning
procedure to construct sub-concepts using examples he or she
provides. The technique was shown to yield more compact models
than an unstructured approach on two large-scale chess problems,
using decision-trees induced for sub-concepts.

Clearly the principal difficulty in the Shapiro-style of structured induction
is the requirement for human intervention at the
structuring step. The following notable efforts in ILP or closely
related areas, have been
directed at learning structured theories automatically:
\begin{itemize}
\item Inverse resolution, especially in the Duce system \cite{mugg:duce}
    was explicitly aimed at learning structured sets of rules in
    propositional logic. The sub-concepts are constructed
    bottom-up;
\item Function decomposition, using HINT \cite{bratko:hint}, which
    learns hierarchies of concepts using automatic top-down
    function decomposition of propositional concepts;
\item First-order theories with exceptions, using the GCWS approach
    \cite{bain:nonmon}, which automatically constructs hiearchical
    concepts. Structuring
    is restricted to learning exceptions to concepts learned at
    a higher level of the hierarchy;
\item First-order tree-learning: an example
    is the TILDE system: \cite{hendrik:tilde}. In this, the
    tree-structure automatically imposes a structuring on the
    models. In addition, if each node in the tree is allowed
    a  ``lookahead''option, then nodes can contain conjunctions
    of first-order literals, each of which can be seen as defining
    a new feature. The model is thus a hierarchy of first-order
    features; and
\item Meta-interpretive learning \cite{mugg:meta_predinv}, which allows
	a very general form of predicate-invention, by allowing an abduction
    step when employing a meta-interpreter to use higher-order templates
    of rules that be used to construct proofs (in effect, explanations) for
    data. In principle, this would allow us not just to construct
    explanations on-demand, but also invent features on-demand. If the
    higher-order templates can be specialised to the domain, then it
    should be possible to control the feature-invention by
    relevance-information. Of course, this is unrelated to
    generating explanations for the predictions of a black-box
    classifier.
\end{itemize}

\noindent
An entirely different, and much more sophisticated kind of hybrid model
combining connectionist and logical components has been proposed recently
in the form of Lifted Relational Neural Networks (LRNNs: \cite{filip:lrnn}). In this, the
logical component is used to provide a template for ground neural network
models, which are used to learn weights on the logical formulae. While we
have largely stayed within the confines of classical ILP both for obtaining
features and explanations, LRNNs are closely related to probabilistic models for
ILP. An area of common interest arises though in the use of the network
structure to invent new features (although in the LRNN case, this is not for
local models as we proposed here).

\section{Concluding Remarks}
\label{sec:concl}

The recent successes of deep neural networks on predictive tasks have not,
to any large extent, used either domain knowledge or representations significantly
more expressive than simple relations (usually over sequences). The price for this has
been a requirement for very large amounts of data, which provide the network with sufficient
correlations necessary to identify predictively useful models. This works for problems where large amounts of
data are being routinely generated automatically; and about which there may be little or no
domain knowledge. The situation with scientific data is quite the opposite: data are sparse,
but there is significant domain-knowledge, often built up over decades of painstaking
experimental work. We would like powerful predictive models in such domains, but for this,
the network would need a way of capturing what is known already, in a language that is sufficiently
expressive. The knowledge-rich deep networks (DRMs) we have proposed here is one way forward, and
the results suggest that we can achieve, and often exceed, the levels of predictivity reached by full first-order
learners. It is important to understand also what the experimental evidence presented
does not tell us. It does not tell us, for example, that a  deep network without first-oder features
will not achieve the same performance as those tabulated here. However, it is not immediately apparent
how this conjecture could be tested, since no more data are available for the problems. However it may be
possible to transfer features constructed by a network trained on other problems with
more data.

Our interests in this paper extend beyond prediction. We want to
construct understandable models. For this we start with the approach taken in \cite{lime}
and propose the use of a proxy for the DRM that acts as a readable explanation.
The proxy in this paper is in the form of a symbolic model
for the predictions made by the DRM, constructed using techniques developed in Inductive Logic
Programming (ILP). 
But there are at least three limitations we see arising from using the approach in \cite{lime}.
First, the goal is to generate a readable proxy for the {\em prediction\/} made by a black-box model (for
us, the DRM is the black box). This need not be the same as a readable proxy for the (true-)value of the instance.
Second, readability does not guarantee comprehensibility: in \cite{michie:engconsc2}
examples are shown of readable, but still incomprehensible models.
Thirdly, an
important quality normally required of good explanations, causality, does
not explicitly play a role.
The first issue is inherent to the purpose of the
model, and we have not attempted to change it here. The incorporation
of a semantic prior based on relevance is a first attempt to address directly the second issue, and
indirectly may address the third partially (non-causal explanations should have
low relevance). To construct causal explanations correctly we will need more information
than assigning relevance labels to predicates. We will also need the explanation-generator
to pose counterfactual queries to the black-box, and the black-box to be able answer such
queries with high accuracy. At this point, there is some evidence that symbolic learning could
be adapted to suggest new experiments (see for example \cite{ross:robot}), but it is not
known how well DRMs will perform if the distribution of input values is very different to
those that were used to train the network. So, at this point, we have restricted ourselves to
constructing readable explanations for predictions that take into account prior preferences.

There are at least three separate directions in which we plan to extend
the work here. First, interactions with the domain-expert suggests that
the relevance information we have used here can be made much more fine-grained.
It is possible, for example that certain combinations of predicates may be more
relevant than others (or, importantly, certain combinations are definitely
not relevant). None of this is accounted for in the current feature-generation
process, and we intend to investigate relevance-guided sampling in place of
the simple random sampling we use at present. Secondly,
we would like to explore the construction of causal explanations for DRMs, by
combining counterfactual reasoning, with the use of a generative deep network
capable of generating new instances. Thirdly, it is necessary at some point
in the future, to establish the link between local symbolic explanations and
human comprehensibility. For this, we would need to conduct a cross-comparison
of structured and unstructured explanations and ratings of their comprehensibility
by a domain-expert. Recently (\cite{mugg:compre}) experiments have been
reported in the ILP literature on assessing comprehensibility,
when invention of predicates is allowed. Similar experiments
will help assess the human-comprehensibility of local symbolic explanations for
black-box classifiers.

\subsection*{Acknowledgements}
A.S. is a Visiting Professorial Fellow, School of CSE, UNSW Sydney.
A.S. is supported by the SERB grant EMR/2016/002766. 

\bibliography{prefs}

\appendix
\label{appa}

\section{Fidelity and Likelihood}
\label{app:lik}

The model for noisy data in \cite{eric:lime} can be
adapted to the construction of local explanations that are not completely consistent with
local predictions (that is, fidelity $<$ 1).
 
\begin{myremark}
{\bf (Bayesian Posterior~\cite{eric:lime})}
\label{def:bayes}
Given a relational instance $a \in {\cal X}$, let $N$ be predictive
model s.t. $N(FV(a)) = c$ for $c \in {\cal Y}$.
Let sets $E^{+}$ and $E^{-}$ denote the local neighbourhood of $a$
and $D=(E^{+},E^{-})$. Let $H$ be a local explanation for $a$ (not necessarily
consistent), using
background knowledge $B$. Then, if
$\theta(H)$ is the proportion of all instances in ${\cal X} \times {\cal Y}$ covered by $H$,
and $\epsilon$ is an estimate of inconsistency (noise)
allowed $\epsilon \in [0,1]$), the log posterior is given by \cite{eric:lime}:
\[
\mathrm{log}P(H|D,B) ~=~ \mathrm{log}P(D|H,B) + \mathrm{log}P(H|B)  - \mathrm{log}(D|B)
\]

where $P(H|B)$ denotes the prior probability, and:
\[
\mathrm{log}P(D|H,B) ~=~ |TP(H)|\mathrm{log}\left(\frac{1-\epsilon}{\theta(H)} + \epsilon \right)
                + |TN(H)| \mathrm{log}\left(\frac{1-\epsilon}{1-\theta(H)} + \epsilon\right) + |FPN(H)|\mathrm{log}(\epsilon) 
\]

\noindent
is the log-likelihood. Here $TP(H)$ is the set $\{e: e \in E^{+} ~\mathrm{and}~B \wedge H \models e\}$;
$TN(H)$ is the set $\{e: e \in E^{-}~\mathrm{and}~B \wedge H \wedge \neg e \not \models \Box\}$; and
$FPN(H)$ = $D - (TP(H) \cup TN(H))$.
\end{myremark}

\noindent
It is not hard to see that $TP$ and $TN$ correspond
to $AgreePos$ and $AgreeNeg$ in Definition \ref{def:fidelity}. In some cases,
it is in fact sufficient
to maximise fidelity, to maximise the log-likelihood.

\begin{myremark}
{\bf (Fidelity and Log-Likelihood)}
\label{rem:fid_loglik}
Let $H_{1,2}$ be local explanations for a relational
example $Class(a,c)$, given $D,B$.
If $\theta(H_1) = \theta(H_2)$,
$|TP(H_2)| \geq |TP(H_1)|$ and $|TN(H_2)| \geq |TN(H_1)|$ then:
(a) $Fidelity(H_2|D,B) \geq Fidelity(H_1|D,B)$; and
(b) $\mathrm{log}P(D|H_2,B) \geq \mathrm{log}P(D|H_1,B)$.\\

Given $D,B$, the log-likelihood $\mathrm{log}~P(D|H,B)$,
is given in Defn.~\ref{def:bayes}. 
From Defn.~\ref{def:fidelity} $Fidelity(H_1|D,B) = \frac{|TP(H_1)|+|TN(H_1)|}{|D|}$ 
and $Fidelity(H_2|D,B) = \frac{|TP(H_2)|+|TN(H_2)|}{|D|}$.
Since $|TP(H_2) \geq |TP(H_1)|$ and $|TN(H_2)| \geq |TN(H_1)|$,
it follows trivially that $Fidelity(H_2|D,B) \geq Fidelity(H_1|D,B)$.

For any explanation $H$, s.t. $0 < \theta)(H) < 1$ and 
For a fixed $\epsilon$ s.t. $0 \leq \epsilon \leq 1$,
the log-multipliers $k_{1,2}(\theta(H))$ of $|TP(H)|$ and $|TN(H)|$ in
the expression for $\mathrm{log}~P(D|H,B)$
are both positive; and the log-multiplier $k_3$ of $|FPN(H)|$ is at most $0$. 
Therefore $\mathrm{log} P(D|H)$ is
$k_1(\theta(H)) |TP| + k_2(\theta(H)) |TN| - k_3(|D|) + k_3(|TP|+|TN|)$.. 
If $\theta(H_1) = \theta(H_2)$, then $k_1(\theta(H_1)) = k_1(\theta(H_2))$ = $k_1$, say
and $k_2(\theta(H_1) = k_2(\theta(H_2))$ = $k_2$, say.
Then $\mathrm{log} P(D|H_1)$ is
$k_1 |TP(H_1)| + k_2 |TN(H_1)| - k_3(|D|) + k_3(|TP(H_1)|+|TN(H_1)|)$
and $\mathrm{log} P(D|H_2)$ is
$k_1 |TP(H_2)| + k_2 |TN(H_2)| - k_3(|D|) + k_3(|TP(H_2)|+|TN(H_2)|)$,
Since $|TP(H_2) \geq |TP(H_1)|$ and $|TN(H_2)| \geq |TN(H_1)|$,
and $k_{1,2} > 0$ and $k_3 \geq 0$
it follows trivially that
$\mathrm{log}P(D|H_2,B) \geq \mathrm{log}P(D|H_1,B)$.
\end{myremark}



\section{Relevance Information}
\label{app:relev}

The following problem-specific relevance assignments for predicates in the
background knowledge were obtained from R.D. King, University
of Manchester. 

\begin{center}
{\small{
\begin{tabular}{|l|c|l|} \hline
Problem     & Relevance & Predicates \\ \hline
Mut188      & 1         & Atoms and bonds \\
          & 2         & 3-dimensional distance \\
                    & 3         & Functional groups and rings \\
                    & 4         & LUMO, hydrophobicity \\
                    & 5         & Expert-identified indicator variables \\ \hline
Canc330     & 1         & Atoms and bonds \\
                    & 2         & Functional groups and rings \\
                  & 3         & Carcinogenic alerts \\
                    & 4         & Outcome of genetic tests \\ \hline
DssTox      &   1         &  Atoms and bonds  \\ \hline
Alzh.       & 1         & Substitutions at templates      \\        
Datasets    & 2         & Hansch-type predicates (size, polarity {\em etc\/}.)\\ \hline
\end{tabular}
}}
\end{center}

\section{Example Explanations}
\label{app:expl}

The following explanations are for a test-instance in the $Canc330$ problem (specifically,
test-instance 2 on the 3rd cross-validation split). This example was chosen since it illustrates a number of interesting aspects: all explanations have perfect fidelity; structuring increases relevance; and there are several structured explanations possible. The DRM has 
2196 features, of which 397 are active for this instance. The DRM correctly predicts the instance as belonging to the ``positive'' class. All the symbolic explanations below predict the same class-values as the DRM for the test-instance and its neighbours.

\begin{tabular}{l} \\
\underline{Unstructured explanation:}\\[6pt]
{Label:} $\langle L_H = 1.0, P_H = \{[1,4]\} \rangle$ \\ 
{Explanation $H$:} \\
\hspace*{0.1cm} $Class(x,c) \leftarrow F_{537}(x), F_{1196}(x), F_{610}(x), F_{611}(x), F_{1657}(x)$ \\[12pt]
\underline{Structured explanation(s): }\\[6pt]
{Label:} $\langle\ L_{H_1} = 1.0, P_{H_1} = \{[1,4],[4,4]\}\rangle$ \\
{Explanation $H_1$:} \\
\hspace*{0.1cm} $Class(x,c) \leftarrow F_{1,1}(x), F_{1,2}(x)$ \\
\hspace*{0.1cm} $F_{1,1}(x) \leftarrow F_{1657}(x)$ \\
\hspace*{0.1cm} $F_{1,2} \leftarrow F_{537}(x), F_{1196}(x), F_{610}(x), F_{611}(x)$ \\[6pt]

{Label:} $\langle\ L_{H_2} = 1.0, P_{H_2} = \{[1,4],[4,4]\}\rangle$ \\
{Explanation $H_2$:} \\
\hspace*{0.1cm} $Class(x,c) \leftarrow F_{1,1}(x), F_{1,2}(x)$ \\
\hspace*{0.1cm} $F_{1,1}(x) \leftarrow F_{611}(x)$ \\
\hspace*{0.1cm} $F_{1,2} \leftarrow F_{537}(x), F_{1196}(x), F_{610}(x), F_{1657}(x)$ \\[6pt]

{Label:} $\langle\ L_{H_3} = 1.0, P_{H_3} = \{[1,4],[4,4]\}\rangle$\\
{Explanation $H_3$:} \\
\hspace*{0.1cm} $Class(x,c) \leftarrow F_{1,1}(x), F_{1,2}(x)$ \\
\hspace*{0.1cm} $F_{1,1}(x) \leftarrow F_{1657}(x), F_{611}(x)$ \\
\hspace*{0.1cm} $F_{1,2} \leftarrow F_{537}(x), F_{1196}(x), F_{610}(x)$ \\[12pt]

\underline{Feature-definitions:}\\
$F_{537}(x) \leftarrow 	Atm(x,y,h,3,z), Gteq(z,0.115)$  \hspace*{0.1cm} (Relev = [1,1]) \\
$F_{1196}(x) \leftarrow Atm(x,y,c,22,z), Gteq(z,-0.111), Atm(x,w,c,22,z)$ \hspace*{0.1cm} (Relev = [1,1]) \\
$F_{610}(x) \leftarrow 	Non\_ar\_hetero\_6\_ring(x,u), Has\_property(x,ames,p)$	\hspace*{0.1cm} (Relev = [2,4]) \\
$F_{611}(x) \leftarrow 	Has\_property(x,salmonella,n),Has\_property(x,mouse\_lymph,p)$ \hspace*{0.1cm} (Relev = [4,4])\\
$F_{1657}(x) \leftarrow Has\_property(x,cytogen\_ca,n), Has\_property(x,mouse\_lymph,p), Has_property(x,cytogen\_sce,p)$ \\ \hspace*{2cm} (Relev = [4,4])
\end{tabular}

\end{document}